\documentclass[10pt,journal,compsoc]{IEEEtran}

\usepackage{cite}

\ifCLASSINFOpdf
\else
\fi

\usepackage{times}
\usepackage{epsfig}
\usepackage{graphicx}
\usepackage{amsmath}
\usepackage{amssymb}

\usepackage{subfig}
\usepackage{booktabs}
\usepackage{multirow}
\usepackage{url}
\usepackage{enumitem}
\usepackage{microtype}
\usepackage[font=small,labelfont=bf]{caption}
\usepackage{xcolor}
\usepackage{diagbox}
\usepackage{nameref}
\usepackage{makecell}
\usepackage{pifont}
\usepackage{tablefootnote}
\usepackage{footnote}

\newcommand{\norm}[1]{\left\lVert#1\right\rVert}
\newcommand{\Paragraph}[1]{\vspace{2mm} \noindent \textbf{#1.} \hspace{0mm}}

\def\L{\mathcal{L}}
\def\Enc{\mathcal{E}}
\def\Dec{\mathcal{D}}
\def\Lxrecon{\L_{\text{xrecon}}}
\def\Lpose{\L_{\text{pose-sim}}}
\def\Lcano{\L_{\text{cano-sim}}}
\def\Lid{\L_{\text{id-inc-avg}}}

\def\fap{\mathbf{f}_a}
\def\fca{\mathbf{f}_{c}}
\def\fpo{\mathbf{f}_{p}}

\def\fdyngait{\mathbf{f}_{\text{dyn-gait}}}
\def\fstagait{\mathbf{f}_{\text{sta-gait}}}

\begin{document}

    \title{On Learning Disentangled Representations \\ for Gait Recognition}

    \author{Ziyuan~Zhang,
    Luan~Tran,
    Feng~Liu,~\IEEEmembership{Member,~IEEE},
    and~Xiaoming~Liu,~\IEEEmembership{Member,~IEEE}
    \IEEEcompsocitemizethanks{
    \IEEEcompsocthanksitem  Ziyuan Zhang,  Luan Tran, Feng Liu, and Xiaoming Liu are with the Department of Computer Science and Engineering, Michigan State University.
    \protect\\ E-mail: \{zhang835, tranluan, liufeng6\}@msu.edu, liuxm@cse.msu.edu
    }
    }

    \markboth{IEEE TRANSACTIONS ON PATTERN ANALYSIS AND MACHINE INTELLIGENCE}%
    {Shell \MakeLowercase{\textit{et al.}}: Bare Demo of IEEEtran.cls for Computer Society Journals}

    \IEEEtitleabstractindextext{%
    \begin{abstract}
Gait, the walking pattern of individuals, is one of the important biometrics modalities. Most of the existing gait recognition methods take silhouettes or articulated body models as gait features. These methods suffer from degraded recognition performance when handling confounding variables, such as clothing, carrying and viewing angle. To remedy this issue, we propose a novel AutoEncoder framework, GaitNet, to explicitly disentangle appearance, canonical and pose features from RGB imagery. The LSTM integrates pose features over time as a dynamic gait feature while canonical features are averaged as a static gait feature. Both of them are utilized as classification features. In addition, we collect a Frontal-View Gait (FVG) dataset to focus on gait recognition from frontal-view walking, which is a challenging problem since it contains minimal gait cues compared to other views. FVG also includes other important variations, e.g., walking speed, carrying, and clothing. With extensive experiments on CASIA-B, USF, and FVG datasets, our method demonstrates superior performance to the SOTA quantitatively, the ability of feature disentanglement qualitatively, and promising computational efficiency. We further compare our GaitNet with state-of-the-art face recognition to demonstrate the advantages of gait biometrics identification under certain scenarios, e.g., long distance/lower resolutions, cross viewing angles.
    \end{abstract}

    }

    \maketitle
    \IEEEdisplaynontitleabstractindextext
    \IEEEpeerreviewmaketitle

    \IEEEraisesectionheading{\section{Introduction}\label{sec:introduction}}

    \IEEEPARstart{B}{iometrics} measures people's unique physical and behavioral characteristics to recognize the identity of an individual.
    Gait~\cite{nixon2010human}, the walking pattern of an individual, is one of biometrics modalities besides face, fingerprint, iris, \emph{etc}.
    Gait recognition has the advantage that it can operate at a distance without users' cooperation.
    Also, it is difficult to camouflage.
    Due to these advantages, gait recognition is applicable to many applications such as person identification, criminal investigation, and healthcare.

    As other recognition problems, gait data can usually be captured by five types of sensors~\cite{wan2018survey}, \emph{i.e.},
    RGB camera,
    RGB-D camera~\cite{Wang_2019_CVPR,zou2017robust},
    accelerometer~\cite{zhang2014accelerometer},  
    floor sensor~\cite{middleton2005floor},  
    and continuous-wave radar~\cite{wang2016gait}. 
    Among them, RGB camera is not only the most popular one due to the low sensor cost, but also the most challenging one since RGB pixels might not be effective in capturing the motion cues.
    This work studies gait recognition from RGB cameras.

    The core of gait recognition lies in extracting {\it gait features} from the video frames of a walking person, where the prior work can be categorized into two types: appearance-based and model-based methods.
    The appearance-based methods, \emph{e.g.}, Gait Energy Image (GEI)~\cite{han2006individual}, take the averaged silhouette image as the gait feature.
    While having a low computational cost and being able to handle low-resolution imagery, it can be sensitive to variations such as cloth change, carrying, viewing angles and walking speed~\cite{sarkar2005humanid,bashir2009gait,wu2017comprehensive,hossain2010clothing,alotaibi2017improved,shiraga2016geinet,Yu2019GaitGANv2IG}.
    The model-based methods use the articulated body skeleton from pose estimation as the gait feature.
    They show more robustness to aforementioned variations but at a price of a higher computational cost and dependency on pose estimation accuracy~\cite{ariyanto2012marionette,choi2019skeleton,feng2016learning}.




    \begin{figure*}%
        \centering
        \subfloat[]{\includegraphics[width=6cm]{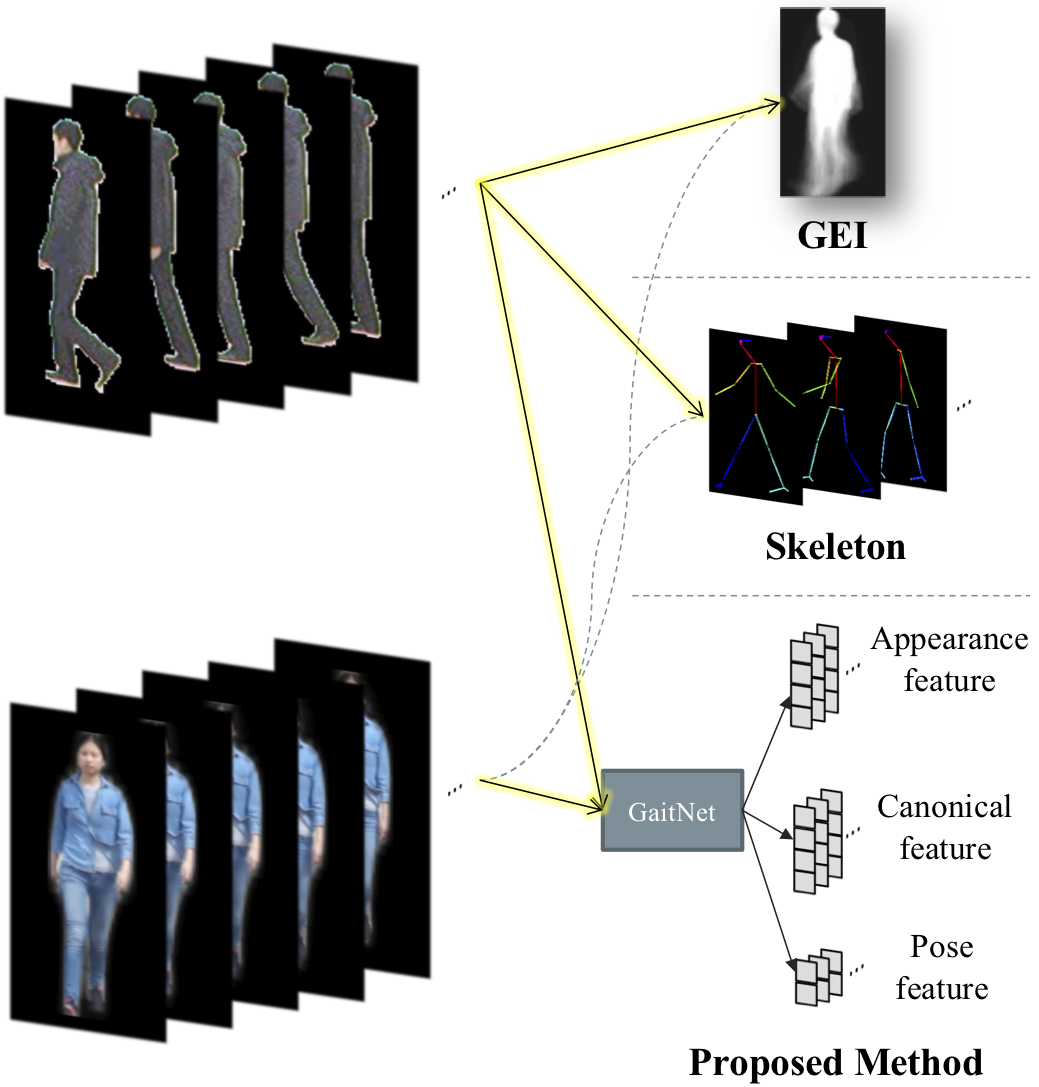}}        \unskip\ \vrule\
        \subfloat[]{\includegraphics[height=6cm]{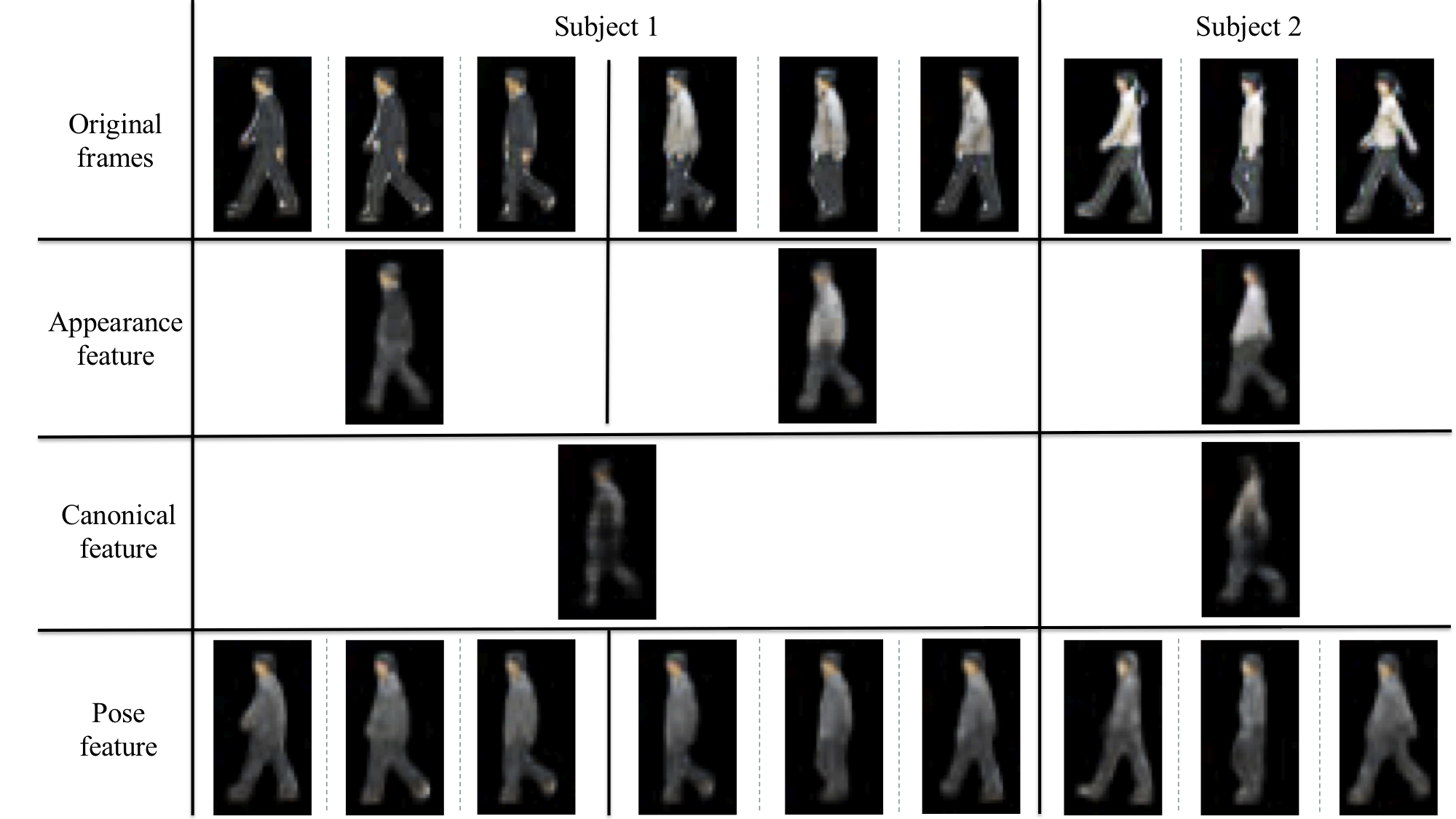}}%

        \caption{(a) While conventional gait databases capture side-view imagery, we collect a new gait database (FVG) with focus on more challenging frontal views. We propose a novel CNN-based model, termed GaitNet, to directly learn the disentangled appearance, canonical and pose features from walking videos, as opposed to handcrafted GEI or skeleton features. (b) Given $2$ videos of Subject $1$ and $1$ video of Subject $2$, feature visualizations by our decoder in Fig.~\ref{fig:arc} show that, the appearance feature is video-specific capturing clothing information; the canonical feature is subject-specific capturing the overall body shape at a standard pose; the pose feature is frame-specific capturing body poses at individual frames.
        }%
        \label{fig:concept}
    \end{figure*}

    %

    It is understandable that the challenge in designing a gait feature is the necessity of being {\it invariant} to the appearance variation due to clothing, viewing angle, carrying, \emph{etc}.
    Therefore, our desire is to {\it disentangle} the gait feature from the non-gait-related appearance of the walking person.
    For both appearance-based or model-based methods, such disentanglement is achieved by manually handcrafting the GEI-like~\cite{han2006individual,bashir2009gait} or body skeleton-like~\cite{ariyanto2012marionette,feng2016learning,choi2019skeleton} features, since neither has color or texture information.
    However, we argue that these manual disentanglements may be sensitive to changes in walking condition.
    In other words, they can lose certain or create redundant gait information.
    E.g., GEI-like features have distinct silhouettes for the same subject wearing different clothes.
    For skeleton-like features, when carrying accessories (\emph{e.g.}, bags, umbrella), certain body joints such as hands may have fixed positions, and hence are redundant information to gait.

    To remedy the aforementioned issues in handcrafted features, as shown in Fig.~\ref{fig:concept}~(a), this paper proposes a novel approach to learn gait representations from the RGB video directly.
    Specifically, we aim to automatically disentangle dynamic pose features (trajectory of gait) from pose-irrelevant features.
    To further distill identity information from pose-irrelevant features, we disentangle the pose-irrelevant features into appearance (\emph{i.e.,} clothing) and canonical features. Here, the canonical feature refers to a standard and unique representation of human body, such as body ratio, width and limb lengths, \emph{etc}.
    The pose features and canonical features are discriminative in identity and are used for gait recognition.
    Fig.~\ref{fig:concept}~(b) visualizes the three disentangled  features.


    This disentanglement is realized by designing an autoencoder-based Convolutional Neural Network (CNN), GaitNet, with novel loss functions.
    For each video frame, the encoder estimates three latent representations: pose, canonical and appearance features,
    by employing three loss functions: 1) cross reconstruction loss enforces that the canonical and appearance features of one frame, fused with the pose feature of another frame, can be decoded to the latter frame;
    2) pose similarity loss forces a sequence of pose features extracted from a video sequence, of the same subject to be similar even under different conditions;
    3) canonical consistency loss favors consistent canonical features among videos of the same subject under different conditions.
    Finally, the pose features of a sequence are fed into a multi-layer LSTM with our designed incremental identity loss to generate the sequence-based dynamic gait feature.
    The average of canonical features results in the sequence-based static gait feature.
    Given two gait videos, the cosine distances between their respective dynamic and static gait features are computed and their summation is the final video-to-video gait similarity metric.

    In addition, most prior work~\cite{han2006individual, bashir2009gait, ariyanto2012marionette, bobick2001gait, cunado2003automatic, Zhang_2019_CVPR, Makihara_2018_CVPR_Workshops, shiraga2016geinet, tao2007general, makihara2017joint} choose the walking video of the side view, which has the richest gait information, as the gallery sequence.
    However, in practices other viewing angles, such as the frontal view, can be very common when pedestrians walk toward or away from the surveillance camera.
    Also, the prior work~\cite{sivapalan2011gait,chattopadhyay2014pose,chattopadhyay2014frontal,nambiar2012frontal} that focuses on frontal view are often based on RGB-D videos, which have additional depth information than RGB.
    Therefore, to encourage gait recognition from frontal-view RGB videos that generally has the minimal amount of gait information, we collect a high-definition (HD, $1080$p) Frontal-View Gait database, named FVG, with a wide range of variations.
    It has three frontal-view angles where the subject walks from left $45^{\circ}$, $0^{\circ}$, and right $45^{\circ}$  off the optical axes of the camera.
    For each of three angles, different variants are explicitly captured including walking speed, clothing, carrying, multiple people, \emph{etc}.


    A preliminary version of this work was published in the IEEE/CVF Conference on Computer Vision and Pattern Recognition (CVPR) $2019$~\cite{gaitnet_cvpr}. We extend the work from three aspects.
    1) Instead of disentangling features in two components: pose and pose-irrelevant~\cite{gaitnet_cvpr}, we further decouple the pose-irrelevant features into discriminative canonical feature and appearance feature. By devising an effective canonical consistency loss, the canonical feature helps to improve gait recognition accuracy.
    2) We conduct more insightful ablation studies to analyze the relationship between our disentanglement losses and features, gait recognition over time, and contributions of dynamic and static gait features.
    3) We perform side-by-side comparison between gait recognition and the state-of-the-art (SOTA) face recognition on the same dataset. 

    In summary, this paper makes the following contributions:

    $\diamond$ Our proposed GaitNet directly learns disentangled representations from RGB videos, which is in sharp contrast to the conventional appearance-based or model-based methods.

    $\diamond$ We introduce a Frontal-View Gait database, including various variations of viewing angles, walking speeds, carrying, clothing changes, background and time gaps. This is the first HD gait database, with nearly twice the number of subjects compared to existing RGB gait databases. 

    $\diamond$ Our proposed method outperforms the state of the arts on three benchmarks, CASIA-B, USF, and FVG datasets.

    $\diamond$ We demonstrate the strength of gait recognition over face recognition in the task of person recognition from surveillance-quality videos.

    \section{Related Work}
    \label{sec:relatedwork}

    \begin{table*}[t!]
        \centering
        \caption{Comparison of existing gait databases and our collected FVG database.}
        \label{tab:db}
        \resizebox{1\linewidth}{!}{

        \begin{tabular}{lccccccl}
            \toprule
            Dataset & \#Subjects & \#Videos & Environment & FPS & Resolution & Format & Variations \\ \midrule
            CASIA-B~\cite{yu2006framework} & $124$ & $13,640$ & Indoor & $25$ & $320{\times}240$  & RGB & \begin{tabular}[c]{@{}c@{}}
                                                                                                             View, Clothing, Carrying
            \end{tabular} \\ 
            USF~\cite{sarkar2005humanid} & $122$ & $1,870$ & Outdoor & $30$ & $720{\times}480$  & RGB & \begin{tabular}[c]{@{}c@{}}
                                                                                                            View, Ground Surface, Shoes, Carrying, Time
            \end{tabular} \\ 
            OU-ISIR-LP~\cite{iwama2012isir} & $4,007$ & - & Indoor & - & $640{\times}480$  & Silhouette & \begin{tabular}[c]{@{}c@{}}
                                                                                                              View
            \end{tabular} \\ 
            OU-ISIR-LP-Bag~\cite{makihara2012isir} & $62,528$ & - & Indoor & - & $1,280{\times}980$  & Silhouette & \begin{tabular}[c]{@{}c@{}}
                                                                                                                        Carrying
            \end{tabular} \\ 
            FVG (ours) & $226$ & $2,856$ & Outdoor & $15$ &  $1,920{\times}1,080$  & RGB & \begin{tabular}[c]{@{}c@{}}
                                                                                               View, Walking Speed, Carrying, Clothing, Multiple people, Time
            \end{tabular} \\ \bottomrule
        \end{tabular}
        }
    \end{table*}



    \Paragraph{Gait Representation}
    Most prior works are based on two types of gait representations.
    In appearance-based methods, gait energy image (GEI)~\cite{han2006individual} or gait entropy image (GEnI)~\cite{bashir2009gait} are defined by extracting silhouette masks.
    Specifically, GEI uses an averaged silhouette image as the gait representation for a video.
    These methods are popular in the gait recognition community for their simplicity and effectiveness. However, they often suffer from sizeable intra-subject appearance changes due to covariates such as clothing, carrying, views, and walking speed.
    On the other hand, model-based methods~\cite{choi2019skeleton,feng2016learning} fit articulated body models to images and extract kinematic features such as $2$D body joints.
    While they are robust to some covariates such as clothing and speed, they require a relatively higher image resolution for reliable pose estimation and higher computational costs.


    In contrast, our approach learns gait representation directly from raw RGB video frames which contain richer information, thus with higher potential of extracting more discriminative gait features.
    The most relevant work to ours is~\cite{chen2018multi}, which learns gait features from RGB images via Conditional Random Field.
    Compared to ~\cite{chen2018multi}, our proposed approach learns two complimentary features: dynamic gait, and static gait features, 
    and has the advantage of being able to leverage a large amount of training data and learning more discriminative representation from data with multiple covariates.
    In addition, some recent works~\cite{wu2017comprehensive,Zhang_2019_CVPR,zhang2019comprehensive, Makihara_2018_CVPR_Workshops, Yu2019GaitGANv2IG} use CNN to learn more discriminative features from GEI.
    However, the source of the learning, GEI, already loses dynamic information since a random shuffle of video frames results in the {\it identical} GEI feature.
    In contrast, the proposed GaitNet learns features from RGB imagery instead, which allows the network to explore richer information for representation learning.
    This is demonstrated by our comparison with~\cite{chen2018multi,wu2017comprehensive} in Sec.~\ref{sec:CASIA-B} and Sec.~\ref{sec:fvg}.

    \Paragraph{Gait Databases}
    There are many classic gait databases such as SOTON Large dataset~\cite{shutler2004large}, USF~\cite{sarkar2005humanid}, CASIA-B~\cite{yu2006framework}, OU-ISIR~\cite{makihara2012isir}, and TUM GAID~\cite{hofmann2014tum}.
    We compare our FVG database with the widely used ones in Tab.~\ref{tab:db}.
    CASIA-B is a large multi-view gait database with three variations: viewing angle, clothing, and carrying.
    Each subject is captured from $11$ views under three conditions: normal walking (NM), walking in coats (CL) and walking while carrying bags (BG).
    For each view,~$6$,~$2$, and $2$ videos are captured in NM, CL and BG conditions, respectively.
    USF database has $122$ subjects with five variations, totaling $32$ conditions per subject.
    It contains two viewing angles (left and right), two ground surfaces (grass and concrete), shoe change, carrying condition and time.
    While OU-ISIR-LP and OU-ISIR-LP-Bag are large databases, only silhouettes are publicly released in both of them.
    In contrast, our FVG focuses on the frontal view, with $3$ different near frontal-view angles toward the camera, and other variations including walking speed, carrying, clothing, multiple people and time.

    \Paragraph{Disentanglement Learning}
    Besides model-based approaches representing data with semantic latent vectors~\cite{tran2018nonlinear, tran2019on, tran2019towards,feng2018disentangling}, data-driven disentangled representation learning approaches are gaining popularity in the computer vision community.
    DrNet~\cite{denton2017unsupervised} disentangles content and pose vectors with a two-encoders architecture, which removes content information in the pose vector by generative adversarial training.
    The work of~\cite{balakrishnan2018synthesizing} segments foreground masks of body parts by $2$D pose joints via U-Net~\cite{ronneberger2015u} and then transforms body parts to desired motion with adversarial training.
    Similarly,~\cite{esser2018variational} utilizes U-net and Variational Auto Encoder (VAE)~\cite{kingma2013auto} to disentangle an image into appearance and shape.
    DR-GAN~\cite{tran2017disentangled,tran2018representation} achieves SOTA performances on pose-invariant face recognition by explicitly disentangling pose variation with a multi-task GAN~\cite{goodfellow2014generative}.
    Different from~\cite{denton2017unsupervised,balakrishnan2018synthesizing,esser2018variational}, our method has only one encoder to disentangle the three latent features, through the design of novel loss functions without the need for adversarial training.
    Further, pose labels are used in DR-GAN training so as to disentangle identity feature from the pose.
    However, to disentangle pose and appearance features from RGB, there is no pose nor appearance {\it label} to be utilized for our method, since it is nontrivial to define the types of walking pattern or clothes as discrete classes.


    \Paragraph{Gait vs.~Face recognition}
    Both gait and face are popular biometrics modalities, especially in covert identification-at-a-distance applications.
    Hence, it is valuable to understand the pros and cons of each modality if the SOTA gait recognition and face recognition algorithms are deployed.
    Along this direction, most of the prior works focus on the fusion of both modalities and evaluate on relatively small datasets~\cite{shakhnarovich2001integrated, kale2004fusion, zhou2007integrating}.
    In contrast, we conduct comprehensive evaluations using SOTA face and gait recognition algorithms, across various conditions of CASIA-B and FVG databases.
    Further, the performances are measured along the video duration to explore the impact of person-to-camera distances.

    \begin{table}[t]
        \renewcommand\arraystretch{1.2}
        \centering
        \caption{Symbols and notations.}\label{tab:symbol}
        \resizebox{0.75\columnwidth}{!}{%
        \begin{tabular}{c|c|l}
            \toprule
            \textbf{Symbol}  & \textbf{Dim.} & \textbf{Notation} \\
            \hline \hline

            $s$ & scalar & Index of subject\\
            $c$ & scalar & Condition\\
            $t$ & scalar &Time step in a video\\
            $n$ & scalar & Number of frames in a video\\
            $\mathbf{X}^{c}$ & matrices &Gait video under condition $c$\\
            $\mathbf{x}^{c, t}$ & matrix &Frame $t$ of video $\mathbf{X}^{c}$ \\
            $\hat{\mathbf{x}}$ & matrix & Reconstructed frame via $\Dec$\\

            \hline
            $\Enc{}$ & - &Encoder network \\
            $\Dec$   & - & Decoder network \\
            $C^{sg}$ & - & Classifier for $\fca$ \\
            $C^{dg}$ & - & Classifier for $\fdyngait$ \\
            \hline
            $\fpo$ & ${64\times 1}$ & Pose feature\\
            $\fca$ & ${128\times 1}$ & Canonical feature\\
            $\fap$ & ${128\times 1}$ & Appearance feature\\
            $\fdyngait$ & ${256\times 1}$ & Dynamic gait feature\\
            $\fstagait$ & ${128\times 1}$ & Static gait feature\\
            $\mathbf{h}^t$ & ${128\times 1}$ & The output of LSTM at step t\\ \hline
            $\Lxrecon$ & - & Reconstruction loss \\
            $\Lpose$ & - & Pose similarity loss \\
            $\Lcano$ & - & Canonical similarity loss \\
            $\Lid$ & - & Incremental identity loss \\
            \bottomrule
        \end{tabular}
        }
    \end{table}

    \section{Proposed Approach}
    \subsection{Overview}

    Let us start with a simple example. Assuming there are three videos, where videos $1$ and $2$ capture subject A wearing t-shirt and long down coat respectively, and in video $3$ subject B wears the same long down coat as in video $2$. The objective is to design an algorithm, from which the gait features of video $1$ and $2$ are the same, while those of video $2$ and $3$ are different. Clearly, this is a challenging objective, as the long down coat can easily dominate the extracted feature, which would make video $2$ and $3$ to be more similar than $1$ and $2$ in the latent space of gait features. Indeed the core challenge, as well as the objective, of gait recognition is to {\it extract gait features that are discriminative among subjects, but invariant to different confounding factors}, such as viewing angles, walking speeds and changing clothes. Table~\ref{tab:symbol} summarizes the symbol and notation used in this paper.

\begin{figure}[t!]
        \centering
        \vspace{-4.5mm}
        \subfloat[The same subject]{{\includegraphics[width=2.34cm]{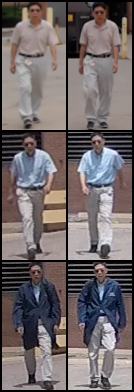} }}%
        \subfloat[Different subjects]{{\includegraphics[width=5.8cm]{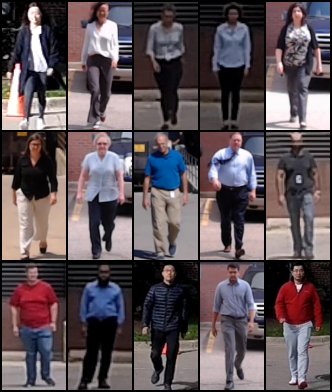} }}%
        \caption{If we may ignore the differences in color/texture of clothing and the body pose, there are inherent body characteristics that are different across subjects (b), and invariant within the same subject (a). These include overall body shape, arm length, torso vs.~leg ratio, \emph{etc}. We define {\it canonical feature} to specifically describe these characteristics.
        }%
        \label{fig:motivation}
    \end{figure}

    Our approach to achieve this objective is feature disentanglement.
    In our preliminary work~\cite{gaitnet_cvpr}, we disentangle features into two components: pose and ``appearance'' features.
    However, further research discovered that the ``appearance'' feature still contains certain discriminative information, which can be useful for identity classification.
    For instance, as in Fig.~\ref{fig:motivation}, imagining if we would ignore the body pose, \emph{e.g.}, position of arms and legs, and clothing information, \emph{e.g.}, color and texture of clothes, we may still tell apart different subjects by their {\it inherent body characteristics}, which can include categories of overall body shape (\emph{e.g.}, rectangle, triangle, inverted triangle, and hourglass~\cite{connell2006body}), arm length, torso vs.~leg ratio~\cite{cheung2003shape}, \emph{etc}.
    In other words, even when different people wearing exactly the same clothing and standing still, these characteristics are still subject dependent.
    In the meantime, for the same subject under various conditions, these characteristics are relatively constant.
    In this work, we term the feature describing these characteristics as the {\it canonical feature}.
    Hence, given a walking video $\mathbf{X}^{c}$ under condition $c$, our framework disentangle the encoded feature into three components: the pose feature $\mathbf{f}_{p}$, the appearance feature $\mathbf{f}_{a}$ and the canonical feature $\mathbf{f}_{c}$.
    We also term the concatenation of  $\mathbf{f}_{a}$ and $\mathbf{f}_{c}$ as the pose-irrelevant feature, which is conceptually equivalent to the ``appearance'' feature in ~\cite{gaitnet_cvpr}.
    The pose feature describes the positions of body parts, and their dynamic over time is essentially the core element of gait;
    the canonical feature defines the unique characteristics of individual body;
    and the appearance feature describes the subject's clothing.

\begin{figure*}[t!]

        \includegraphics[width=0.97\linewidth]{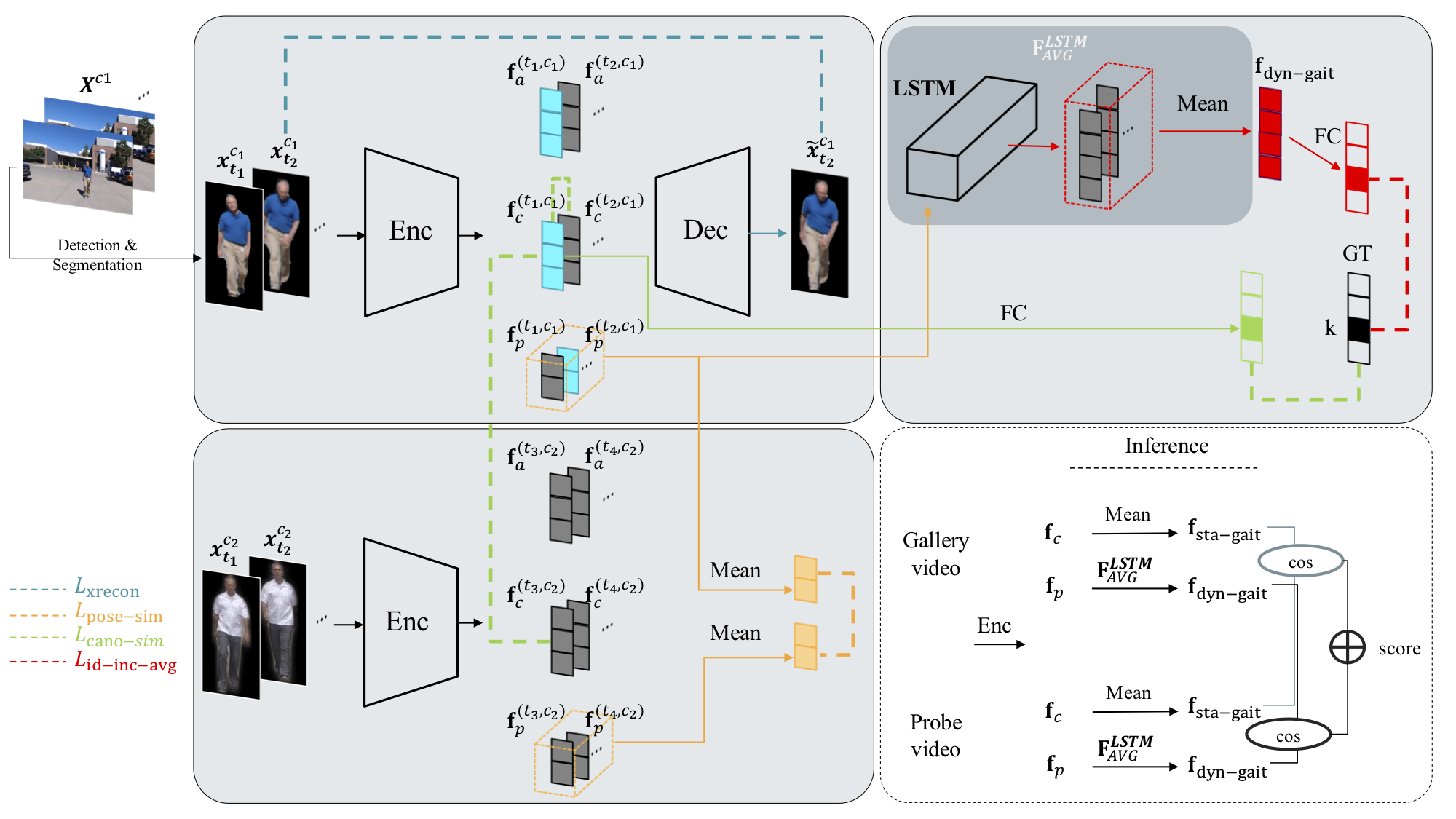}
        \centering
        \caption{The overall architecture of proposed GaitNet. The bottom right block indicates the inference process, while the remaining illustrates the training process with the four color-coded loss functions.}
        \label{fig:arc}
    \end{figure*}

    The above feature disentanglement can be naturally implemented as an encoder-decoder network.
    Specifically, as depicted in Fig.~\ref{fig:arc}, the input to our GaitNet is a video sequence, with background removed using any off-the-shelf pedestrian detection and segmentation method~\cite{he2017mask,illuminating-pedestrians-via-simultaneous-detection-segmentation,pedestrian-detection-with-autoregressive-network-phases}.
    With carefully designed loss functions, an encoder is learned to disentangle the pose, canonical and appearance features for each video frame.
    Then, a multi-layer LSTM explores the temporal dynamics of pose features and aggregates them to a sequence-based dynamic gait feature.
    In the meantime, the average of all the canonical features is defined as the static gait feature.
    Measuring distances of both dynamic and static features between the gallery and probe walking videos provides the final matching score.
    In this section, we first present the feature disentanglement, followed by temporal aggregation, model inference and finally implementation details.


    \subsection{Feature Disentanglement}
    \label{sec:disentangle}

    \begin{table}[t!]
        \renewcommand\arraystretch{1.2}

        \centering
        \caption{ The properties of three disentangled features in terms of its constancy across frames and conditions, and discriminativeness. These properties are the basis for us to design loss functions for feature disentanglement.}
        \label{tab:motivation}
        \resizebox{\linewidth}{!}{
        \begin{tabular}{c|ccc}
            \toprule
            & \textbf{Constant Across Frames}  & \textbf{Constant Across Conditions} & \textbf{Discriminative}\\ \hline \hline
            $\fap$ & Yes & No & No\\ \hline
            $\fca$ & Yes & Yes & Yes \\ \hline
            $\fpo$ & No & Yes & Yes for $\fpo$ over $t$ \\
            \bottomrule
        \end{tabular}
        }
    \end{table}

    For the majority of gait datasets, there is limited intra-subject appearance variation.
    Hence, appearance could be a discriminative cue for identification during training as many subjects can be easily distinguished by their clothes.
    Unfortunately, any feature extractors relying on appearance will not generalize well on the test set or in practice, due to potentially diverse clothing or appearance between two videos of the same subject.
    This limitation on training sets also prevents us from learning ideal feature extractors if solely relying on identification objective.
    Hence we propose to learn to disentangle the canonical and pose feature from the visual appearance.
    Since a video is composed of frames, disentanglement should be conducted at the frame level first.

    Before presenting the details of how we conduct disentanglement, let us first understand the various properties of three types of features, as summarized in Tab.~\ref{tab:motivation}.
    These properties are crucial in guiding us to define effective loss functions for disentanglement.
    The appearance feature mainly describes the clothing information of the subject.
    Hence it is constant within a video sequence, but often different across different conditions. Of course it is not discriminative among individuals.
    The canonical feature is subject-specific, and is therefore constant across both video frames, and conditions.
    The pose feature is obviously different across video frames, but is assumed to be constant across conditions.
    Since the pose feature is the manifestation of video-based gait information at a specific frame, the pose feature itself might not be discriminative.
    However, the dynamics of pose features over time will constitute the dynamic gait feature, which is discriminative among individuals.


    To this end, we propose to use an encoder-decoder network architecture with carefully designed loss functions to disentangle the pose feature and canonical feature from appearance feature.
    The encoder, $\Enc$, encodes a feature representation of each frame, $\mathbf{x}$, and explicitly splits it into three components, namely appearance feature $\fap$, canonical feature $\fca$ and pose feature $\fpo$:
    \begin{equation}
        \label{eq1}
        \fap, \fca, \fpo = \Enc(\mathbf{x}).
    \end{equation}
    Collectively these three features are expected to fully describe the original input image.
    As they can be decoded back to the original input through a decoder $\Dec$:
    \begin{equation}
        \label{eq2}
        \hat{\mathbf{x}} = \Dec(\fap, \fca, \fpo).
    \end{equation}
    We now define the various loss functions to jointly learn the encoder $\Enc$ and decoder $\Dec$.

    \Paragraph{Cross Reconstruction Loss}
    The reconstructed image $\hat{\mathbf{x}}$ should be close to the original input $\mathbf{x}$.
    However, enforcing self-reconstruction loss as in typical auto-encoder cannot ensure the meaningful disentanglement as in our design. 
    Hence, we propose the cross reconstruction loss, using the appearance feature $\fap^{t_1}$ and canonical feature $\fca^{t_1}$ of frame $t_1$ and the pose feature $\fpo^{t_2}$  of frame  $t_2$ to reconstruct the latter frame:
    \begin{equation}
        \label{eq3}
        \Lxrecon = \norm{ \Dec(\fap^{t_2}, \fca^{t_1}, \fpo^{t_1})-\mathbf{x}^{t_2} }^2_2.
    \end{equation}

    The cross reconstruction loss, on one hand, can act as the self-reconstruction loss to make sure the three features are sufficiently representative to reconstruct a video frame.
    On the other hand, as we can pair a pose feature of a current frame with the canonical and appearance features of {\it any} frame in the same video to reconstruct the same target, it enforces both the canonical and appearance features to be similar across all frames within a video.
    Indeed, according to Tab.~\ref{tab:motivation}, between the pose-irrelevant feature, $\fap\&\fca$, and the pose feature $\fpo$, the main distinct property is that the former is constant across frames while the latter is not.
    This is the basis for designing our cross reconstruction loss.

    \Paragraph{Pose Similarity Loss}
    The cross reconstruction loss is able to prevent the pose-irrelevant feature, $\fap\&\fca$, to be contaminated by the pose information that changes across frames.
    If not, \emph{i.e.},  $\fap$ or $\fca$ contains some pose information, $\Dec(\fap^{t_2}, \fca^{t_1}, \fpo^{t_1})$ and $\mathbf{x}^{t_2}$ would have different poses.
    However, clothing/texture and body information may still be leaked into the pose feature $\fpo$.
    In the extreme case, $\fca$ and $\fap$ could be constant vectors while $\fpo$ encodes all the information of a video frame.

    To encourage $\fpo$ including {\it only} the pose information, we leverage multiple videos of the same subject.
    Given two videos of the same subject with length $n_1$, $n_2$ in two different conditions $c_1$, $c_2$, they contain difference in the person's appearance, \emph{i.e.}, cloth changes.
    Despite appearance changes, the gait information is assumed to be constant between two videos.
    Since it's almost impossible to enforce similarity on $\fpo$ between video frames as it requires precise frame-level alignment, we minimize the similarity between two videos' averaged pose features:
    \begin{equation}
        \label{eq4}
        \L_{\text{pose-sim}} =\norm{ \frac{1}{n_1}\sum_{t=1}^{n_1} \fpo^{(t,c_1)} - \frac{1}{n_2}\sum_{t=1}^{n_2} \fpo^{(t,c_2)} }^2_2.
    \end{equation}

    According to Tab.~\ref{tab:motivation}, the pose feature is constant across conditions, which is the basis of our pose similarity loss.

    \Paragraph{Canonical Consistency Loss}
    The canonical feature describes the subject's body characteristics, which is unique over all video frames.
    To be specific, for two videos of the same subject $k$ in two different conditions $c_1$, $c_2$, the canonical feature is constant across both frames and conditions, as illustrated in Tab.~\ref{tab:motivation}.
    Tab.~\ref{tab:motivation} also states that the canonical feature is discriminative across subjects.
    Hence, to enforce the two constancy and the discriminativeness, we define the canonical consistency loss as follows:
    \begin{align}
        \label{eq5}
        \L_{\text{cano-cons}} =
        & \frac{1}{n_1^2} \sum_{i \neq j} \norm{ \fca^{(t_i,c_1)}-\fca^{(t_j,c_1)} }^2_2  \nonumber \\
        & + \frac{1}{n_1} \sum_{i} \norm{ \fca^{(t_i,c_1)}-\fca^{(t_i,c_2)} }^2_2 \nonumber \\
        & + \frac{1}{n_1} \sum_{i} - \log( C_k^{sg}(\fca^{(t1,c1)}) )),
    \end{align}
    where the three terms measure the consistency across frames in a single video, consistency across different videos of the same subject, and identity classification using a classifier $C^{sg}$, respectively.


    \subsection{Gait Feature Learning and Aggregation}
    Even when we can disentangle pose, canonical and appearance information for each video frame, the $\fpo$ and $\fca$ have to be aggregated over time, since 1) gait recognition is conducted between two videos instead of two images; 2) not all the $\fca$ from every single frame is guaranteed to have same canonical information; 3) the current feature $\fpo$ only represents the walking pose of the person at a specific instance, which can share similarity with another instance of a different individual.
    Here, we are looking for discriminative characteristics in a person's walking pattern.
    Therefore, modeling its aggregation for $\fca$ and temporal change for $\fpo$ is critical.

    \subsubsection{Static Gait Feature via Canonical Feature Aggregation}
    After learning $\fca$ for every single frame as defined in Eqn.~\ref{eq5}, we explore the best representation of $\fca$ features across all frames of a video sequence.
    Since $\fca$ is assumed to be constant over time, we compute the averaged $\fca$ features as a way to aggregate the canonical features over time.
    Given that $\fca$ describes the body characteristics as if we freeze the gait, we call the aggregated $\fca$  as the static gait feature $\fstagait$.
    \begin{equation}
        \fstagait = \frac{1}{n}\sum_{t=1}^{n} \fca^{t}.
    \end{equation}

    \subsubsection{Dynamic Gait Feature via Pose Feature Aggregation}
    For temporal modeling of poses, this is where temporal modeling architectures like the recurrent neural network or long short-term memory (LSTM) work best.
    Specifically, in this work, we utilize a multi-layer LSTM structure to explore temporal information of pose features, \emph{e.g.}, how the trajectory of subjects' body parts changes over time.
    As shown in Fig.~\ref{fig:arc}, pose features extracted from one video sequence are fed into a $3$-layer LSTM.
    The output of the LSTM is connected to a classifier $C^{dg}$, in this case, a linear classifier is used, to classify the subject's identity.

    Let $\mathbf{h}^t$ be the output of the LSTM at time step $t$, which is accumulative after feeding $t$ pose features $\fpo$ into it:
    \begin{equation}
        \mathbf{h}^t = \text{LSTM} ( \fpo^1,\fpo^2, ..., \fpo^t ).
    \end{equation}

    Now we define the loss function for LSTM.
    A trivial option for identification is to add the classification loss on top of the LSTM output of the final time step:
    \begin{equation}
        \label{eqn:id_old}
        \L_{\text{id-single}} = - \log( C_k^{dg}( \mathbf{h}^n ) ),
    \end{equation}
    which is the negative log likelihood that the classifier $C^{dg}$ correctly identifies the final output $\mathbf{h}^n$ as its identity label $k$.

    \Paragraph{Identification with Averaged Feature}
    By the nature of LSTM, the output $\mathbf{h}^t$ can be greatly affected by its last input $\fpo^t$.
    Hence the LSTM output, $\mathbf{h}^t$, could be unstable across time steps.
    With a desire to obtain a gait feature that is robust to the final instance of a walking cycle, we choose to use the averaged LSTM output as our gait feature for identification:
    \begin{equation}
        \label{eqn:dyngait}
        \fdyngait^t = \frac{1}{t} \sum_{s=1}^t \mathbf{h}^s.
    \end{equation}

    The identification loss can be rewritten as:
    \begin{align}
        \label{eqn:id_averaged}
        \L_{\text{id-avg}} & = - \log( C_k^{dg}( \mathbf{f}_{\text{dyn-gait}}^n ) ) \nonumber \\
        & = -\log\left( C_k^{dg}\left( \frac{1}{n} \sum_{s=1}^n \mathbf{h}^s \right) \right).
    \end{align}

    \Paragraph{Incremental Identity Loss}
    LSTM is expected to learn that, the longer the video sequence, the more walking information it processes thus the more confident it identifies the subject.
    Instead of minimizing the loss at the final time step, we propose to use all the intermediate outputs of every time step weighted by $w_t$: 
    \begin{equation}
        \label{eqn:id}
        \L_{ \text{id-inc-avg} } = \frac{1}{\sum_{t=1}^{n} w_t }\sum_{t=1}^{n} - w_t \log \left( C_k^{dg} \left( \frac{1}{t} \sum_{s=1}^t \mathbf{h}^s  \right)\right),
    \end{equation}
    where we set $w_t=t^2$ and other options such as $w_t=1$ also yield similar performance.
    In the experiments, we will ablate the impact of three options in classification loss: $\L_{\text{id-single}}$, $\L_{\text{id-avg}}$, and $\L_{ \text{id-inc-avg} }$.
    To this end, the overall loss function is:
    \begin{equation}
        \L = \L_{\text{id-inc-avg}} + \lambda_{r} \Lxrecon + \lambda_d  \L_{\text{pose-sim}} + \lambda_s  \L_{\text{cano-sim}}.
        \label{eqn:finalloss}
    \end{equation}

    The entire system, including encoder, decoder, and LSTM, are jointly trained.
    Updating $\Enc$ to optimize $\L_{\text{id-inc-avg}}$ also helps to further generate pose feature that has identity information and from which LSTM is able to explore temporal dynamics.

    \subsection{Model Inference}
    Since GaitNet takes one video sequence as input and outputs $\fdyngait$ and $\fstagait$ as shown in Fig.~\ref{fig:arc}, one single score is needed to measure the similarity between the gallery and probe videos for either gait authentication or identification.
    During testing, both $\fstagait$ and $\fdyngait$ are used as the identity features for score calculation.
    We use the cosine similarity scores, normalized to the range of $[0,1]$ via min-max.
    The static and dynamic scores are finally fused by a weighted sum rule:
    \begin{align}
        \label{eq_fusion}
        \text{Score} & =  (1-\alpha) * \cos{(\fstagait^g, \fstagait^p)} \nonumber \\
        & + \alpha * \cos{(\fdyngait^g, \fdyngait^p)},
    \end{align}
    where $g$ and $p$ represent gallery and probe, respectively.


    \begin{table}[t]
        \renewcommand\arraystretch{1.2}
        \caption{The architecture of $\Enc$ and $\Dec$ networks. Note the layer with ()* is removed for experiments with small training sets, \emph{i.e.}, all ablation studies in Sec.~\ref{sec:ablation_study}, to prevent overfitting.}
        \label{tab:networkED}
        \centering
        \resizebox{\linewidth}{!}{
        \begin{tabular}{ cccccc  }
            \toprule
            \multicolumn{3}{c}{$\Enc$} & \multicolumn{3}{c}{$\Dec$} \\
            \hline
            Layers & Filters/Stride & Output Size & Layers & Filters/Stride & Output Size \\
            \hline
            Conv1 & $3$x$3$/$1$    &$64$x$32$x$64$& FC & - &   $4$x$2$x$512$\\
            MaxPool1& $3$x$3$/$2$    &$32$x$16$x$64$& FCConv1 &$3$x$3$/$2$&   $8$x$4$x$256$\\
            Conv2 & $3$x$3$/$1$    &$32$x$16$x$256$& FCConv2 &$3$x$3$/$2$&   $16$x$8$x$128$\\
            MaxPool2& $3$x$3$/$2$    &$16$x$8$x$256$& FCConv3 &$3$x$3$/$2$&   $32$x$16$x$64$\\
            Conv3 & $3$x$3$/$2$    &$16$x$8$x$512$& FCConv4 &$3$x$3$/$2$&   $32$x$16$x$3$\\
            (Conv4 & $3$x$3$/$2$    &$16$x$8$x$512$)$^*$ &\\
            MaxPool3 & $3$x$3$/$2$    &$4$x$2$x$512$&\\
            FC & - &$320$&\\
            \bottomrule
        \end{tabular}
        }
    \end{table}

    \subsection{Implementation Details}

    \Paragraph{Detection and Segmentation}
    Our GaitNet receives video frames with the person of interest segmented.
    The foreground mask is obtained from the SOTA instance segmentation algorithm, Mask R-CNN~\cite{he2017mask}.
    Instead of using a zero-one mask by hard thresholding, we maintain the soft mask returned by the network, where each pixel indicates the probability of being a person.
    This is partially due to the difficulty in choosing an appropriate threshold suitable for multiple databases.
    Also, it remedies the loss in information due to the mask estimation error.
    We use a bounding box with a fixed ratio of width : height $= 1: 2$ with the absolute height and center location given by the Mask R-CNN network.
    The input of GaitNet is obtained by pixel-wise multiplication between the mask and the $[0,1]$-normalized RGB values, and then resizing to $32\times64$ pixels.
    This applies to all the experiments on CASIA-B, USF and FVG datasets in Sec.~\ref{sec:experiments}.



    \Paragraph{Network Structure and Hyperparameter}
    Our encoder-decoder network is a typical CNN, illustrated in Tab.~\ref{tab:networkED}.
    Different from our preliminary work~\cite{gaitnet_cvpr}, we replace stride-$2$ convolution layers with stride-$1$ convolution layers and max pooling layers, since we find the latter is able to achieve the similar results with less hyper-parameter searching for different training scenarios.
    Each convolution layer is followed by Batch Normalization and Leaky ReLU activation.
    The decoder structure, similar to~\cite{radford2015unsupervised}, is built from transposed $2$D convolution, Batch Normalization and Leaky ReLU layers.
    The final layer is a Sigmoid activation which can output the value into $[0, 1]$ range as the input.     
    All the transposed convolutions are with stride of $2$ to up sample images and all the Leaky ReLU are with slope of $0.2$.
    The classification part is a stacked $3$-layer LSTM~\cite{gers1999learning}, which has $256$ hidden units in each cell.
    The length of $\fap$, $\fca$ and $\fpo$ is $128$, $128$ and $64$ respectively, as shown in Tab.~\ref{tab:symbol}.

    The Adam optimizer~\cite{kingma2014adam} is initialized with the learning rate of $0.0001$, and the momentum of $0.9$.
    To prevent over-fitting, the weights decay of $0.001$ is applied to all the experiments, and the learning rate decays by multiplying $0.9$ in every $500$ iterations.
    For each batch, we use video frames from $16$ or $32$ different clips depending on different experiment protocols.
    Since video lengths are varied, a random crop of $20$-frame sequence is applied during training; all shorter videos are discarded.
    %
    The $\lambda_{r}$, $\lambda_{s}$ and $\lambda_{d}$ in Eqn.~\ref{eqn:finalloss} are all set to $1$ in all experiments.


    \begin{figure}[t]
        \includegraphics[width=\linewidth]{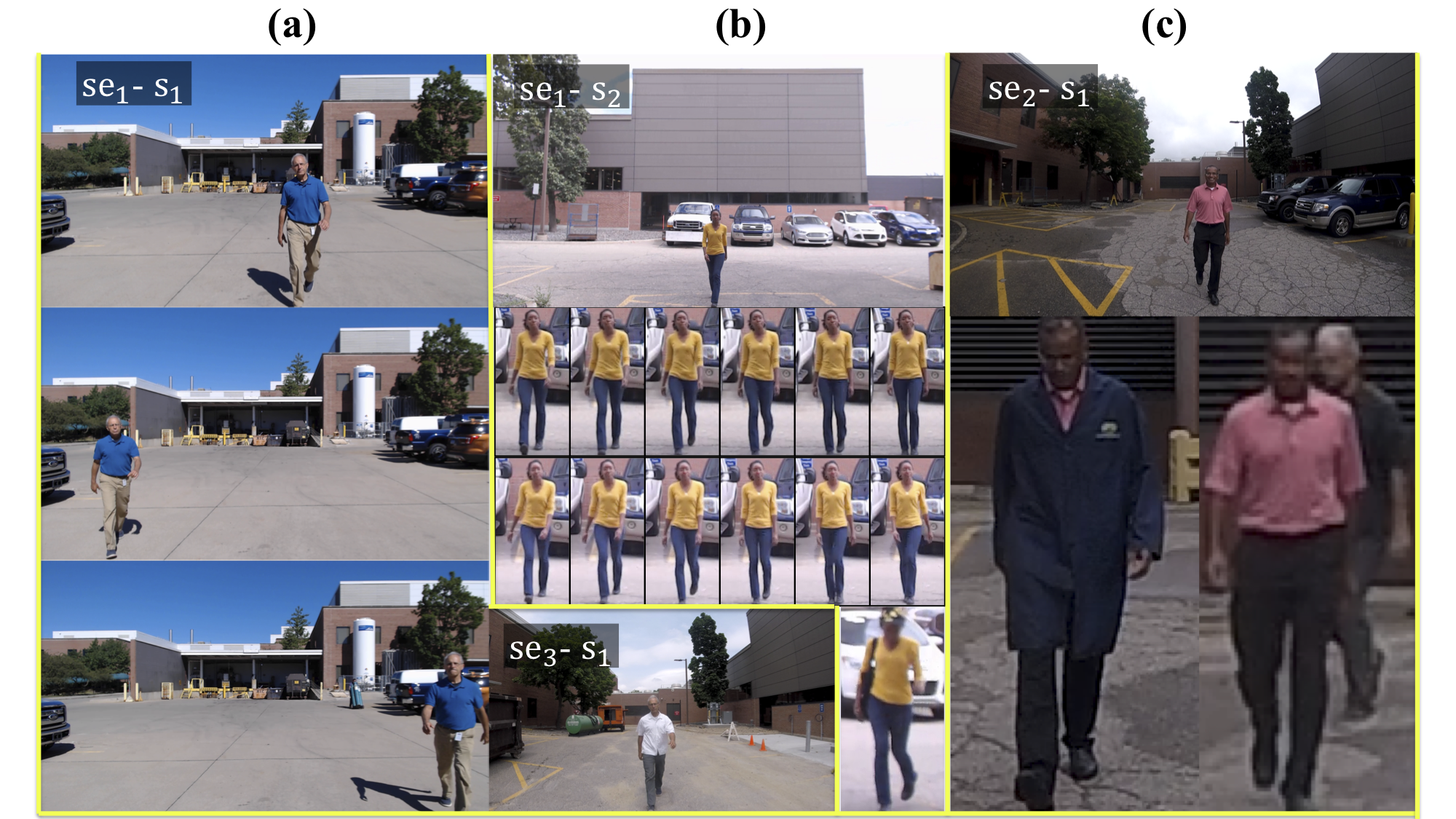}
        \centering
        \caption{Examples of FVG Dataset. (a) Samples of the near frontal middle, left and right walking viewing angles in Session $1$ ($se_1$) of the first subject ($s_1$). $se_3$-$s_1$ is the same subject in Session $3$. (b) Samples of slow and fast walking speed for another subject in Session $1$. Frames in the second row are normal and in the third row are fast walking. Carrying bag and wearing hat sample is shown below. (c) Samples of changing clothes and with multiple people background from one subject in Session $2$. }
        \label{fig:fvg}
    \end{figure}





    \section{Front-View Gait (FVG) Database}
    \label{sec:fvg_intro}

    \Paragraph{Collection}
    To facilitate the research of gait recognition from frontal-view angles, we collect the Front-View Gait (FVG) database in a course of two years ($2017$ and $2018$).
    During the capturing, we place the camera (Logitech C$920$ Pro Webcam or GoPro Hero $5$) on a tripod at the height of $1.50$ meters.
    We require each of $226$ subjects to walk toward the camera $12$ times starting from around $16$ meters away from the camera, which results in $12$ videos per subject.
    The videos are captured at $1,080\times1,920$ resolution with $15$ FPS and the average length of $10$ seconds.
    The height of body in the video ranges from $101$ to $909$ pixels, and the height of faces ranges from $17$ to $467$ pixels.
    These $12$ walks have the combination of three angles toward the camera ($-45^\circ$, $0^\circ$,  $45^\circ$ off the optical axes of the camera), and four variations.
    As detailed in Tab.~\ref{tab:fvg_database}, FVG is collected in three sessions with five variations: normal, walking speed (slow and fast), clothing changes, carrying/wearing change (bag or hat), and clutter background (multiple persons).
    The five variations are well balanced in three sessions. 
    Fig.~\ref{fig:fvg} shows exemplar images from FVG.



    \begin{table}[t!]
        \centering
        \caption{The FVG database. The last $5$ rows show the specific variations that are captured by each of $12$ videos per subject.}
        \label{tab:fvg_database}
        \resizebox{0.48\textwidth}{!}{%
        \begin{tabular}{l|c|c|c|c|c|c|c|c|c}
            \toprule
            Collection Year & \multicolumn{3}{c|}{$2017$} & \multicolumn{6}{c}{$2018$}                      \\ \hline
            Session & \multicolumn{3}{c|}{$1$}      & \multicolumn{3}{c|}{$2$}  & \multicolumn{3}{c}{$3$}  \\ \hline
            Number of Subjects & \multicolumn{3}{c|}{$147$}    & \multicolumn{3}{c|}{$79$} & \multicolumn{3}{c}{$12$} \\ \hline
            Viewing Angle~($^\circ$) & -$45$     & $0$       & $45$     & -$45$     & $0$     & $45$   & -$45$     & $0$     & $45$   \\ \hline
            Normal & $1$       & $2$       & $3$      & $1$      & $2$     & $3$    & $1$       & $2$     & $3$    \\
            Fast / Slow Walking & $4$/$7$     & $5$/$8$     & $6$/$9$    & $4$       & $5$     & $6$    & $4$       & $5$     & $6$    \\
            Carrying Bag / Hat & $10$      & $11$      & $12$     & - & - & - & - & - & -  \\
            Change Clothes & - & - & - & $7$       & $8$     & $9$    & $7$       & $8$     & $9$    \\
            Multiple Person & - & - & - & $10$      & $11$    & $12$   & $10$      & $11$    & $12$   \\
            \bottomrule
        \end{tabular}%
        }
    \end{table}

    \Paragraph{Protocols}
    Different from prior gait databases, subjects in FVG are walking toward the camera, which creates a great challenge on exploiting gait information as the visual difference in consecutive frames is normally much smaller than side-view walking.
    We focus our evaluation on variations that are challenging, \emph{e.g.}, different clothes, carrying a bag while wearing a hat, or are not presented in prior databases, \emph{e.g.}, multi-person. 
    To benchmark research on FVG, we define $5$ evaluation protocols, among which there are two commonalities: $1)$ the first $136$ and remaining $90$ subjects are used for training and testing respectively; $2)$ the video $2$, the normal frontal-view walking, is always used as the gallery.
    The $5$ protocols differ in their respective probe data, which cover the variations of Walking Speed (WS), Carrying Bag while Wearing a Hat (BGHT), Changing Clothes (CL), Multiple Persons (MP), and all variations (ALL).
    At the top part of Tab.~\ref{tab:fvg_database}, we list the detailed probe sets for all $5$ protocols.
    For instance, for the WS protocol, the probes are video $4{-}9$ in Session $1$ and video $4{-}6$ in Session $2$.
    In all protocols, the performance metrics are the True Accept Rate (TAR) at $1\%$ and $5\%$ False Alarm Rate (FAR).

    \section{Experimental Results}
    \label{sec:experiments}

    We evaluate the proposed approach on three gait databases, CASIA-B~\cite{yu2006framework}, USF~\cite{sarkar2005humanid} and FVG.
    As mentioned in Sec.~\ref{sec:relatedwork}, CASIA-B and USF are the most widely used gait databases, which helps us to make the comprehensive comparison with prior works.
    We compare our method with~\cite{wu2017comprehensive,chen2018multi,kusakunniran2010support,kusakunniran2014recognizing} on these two databases, by following the respective experimental protocols of the baselines.
    These are either the most recent and SOTA work, or classic gait recognition methods.
    The OU-ISIR database~\cite{makihara2012isir} is not evaluated, and related results~\cite{makihara2017joint} are not compared since our work consumes RGB video input, but OU-ISIR only releases silhouettes.
    Finally, we also conduct experiments to compare our gait recognition with the state-of-the-art face recognition method ArcFace~\cite{deng2019arcface} on the CASIA-B and FVG datasets.


    \begin{figure}[t]
        \includegraphics[width=9cm]{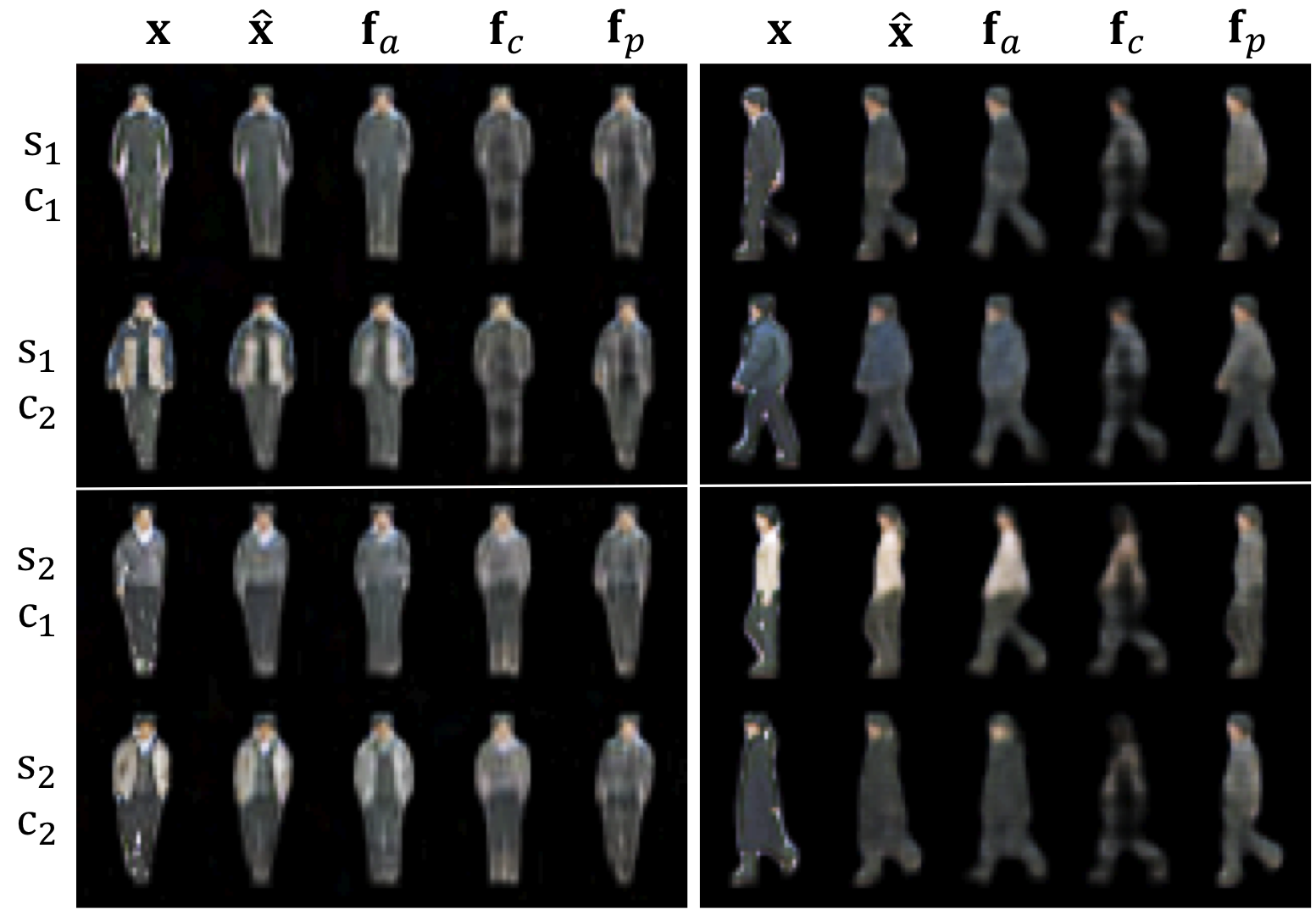}
        \centering
        \caption{
        Synthesis by decoding three features individually, $\fap$, $\fca$ and $\fpo$, and their concatenation.
        Left and right parts are two learnt models on frontal and side views of CASIA-B.
        The top two rows are two frames of the same subject under different conditions (NM vs. CL) and the bottom two are another subject. 
        The reconstructed frames $\hat{\mathbf{x}}$ closely match the original input.
        $\fca$ shows consistent body shape for the same subject while different for different subjects.
        $\fap$ recovers the appearance of clothes, at the pose specified by $\fca$.
        The body pose of $\fpo$ matches with the input frame.
        }
        \label{fig:decodeimages_single}
    \end{figure}

    \begin{figure}[t]
        \includegraphics[width=9cm]{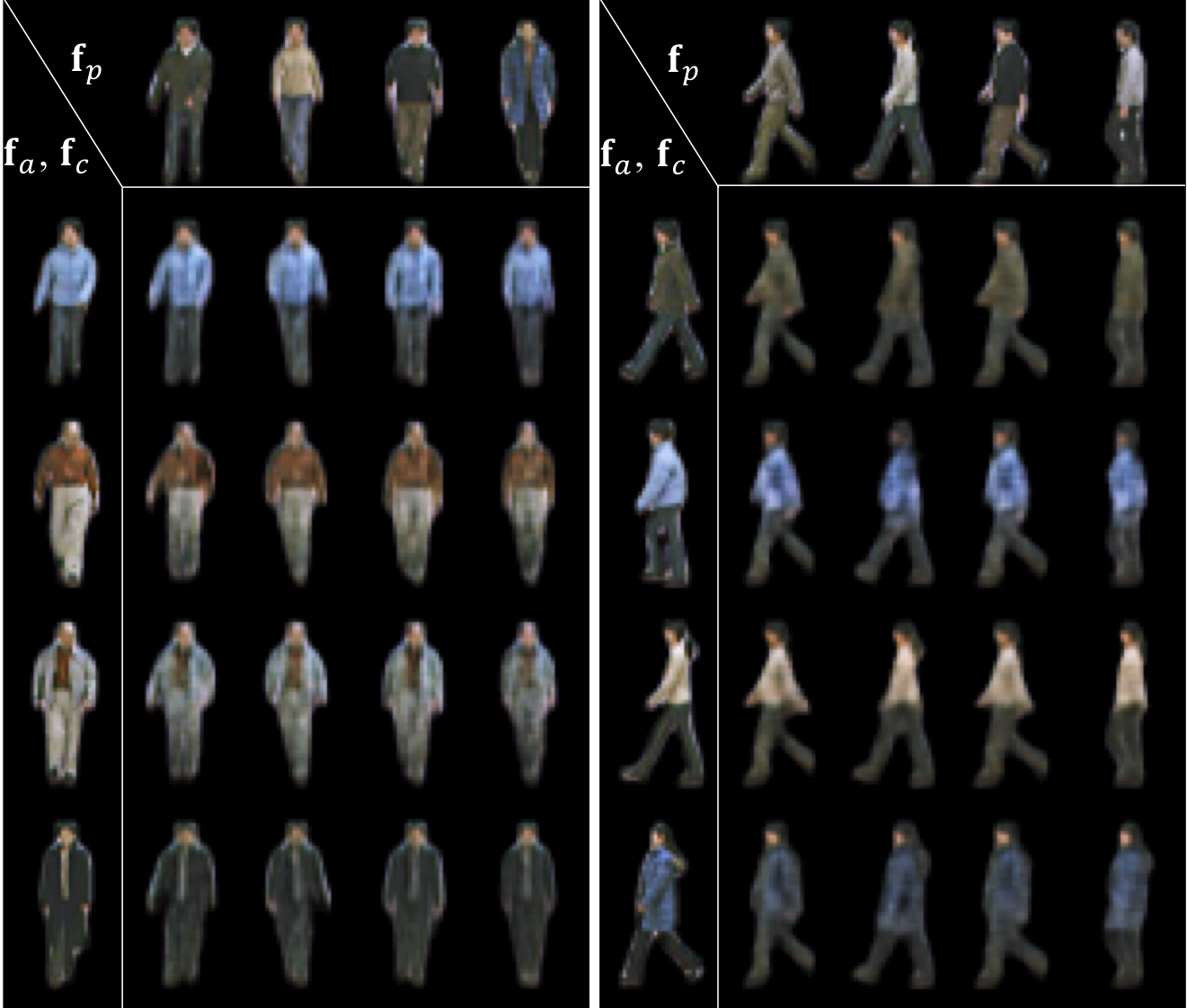}
        \centering
        \caption{
        Synthesis by decoding pairs of pose features $\fpo$ and pose-irrelevant features, $\{\fap, \fca\}$.
        Left and right parts are examples of frontal and side views of CASIA-B.
        In either part, each of $4\times4$ synthetic images is $\Dec(\fap^l, \fca^l, \fpo^t)$, where $\{\fap^l, \fca^l\}$ is extracted from images in the first column and $\fpo^t$ is from the top row.
        The synthetic images resemble the appearance of the first column and the pose of the top row. 
        }
        \label{fig:decodeimages_cross}
    \end{figure}


    \subsection{Ablation Study}
    \label{sec:ablation_study}

    \subsubsection{Feature Visualization Through Synthesis}
    \label{sec:fv_syn}
    While our decoder is only useful in training, but not model inference, it can enable us to visualize the disentangled features as a synthetic image, by feeding either the feature itself, or their random concatenation, to our learned decoder $\Dec$.
    This synthesis helps to gain more understanding of the feature disentanglement.

    \Paragraph{Visualization of Features in One Frame}
    Our decoder requires the concatenation of three vectors for synthesis.
    Hence, to visualize each individual feature, we concatenate it with two vectors of zeros and then feed to decoder.
    In Fig.~\ref{fig:decodeimages_single}, we show the disentanglement visualization of $4$ subjects (two frontal and two side views), each under the NM and CL conditions.
    First of all, the canonical feature discovers a {\it standard} body pose that is consistent across both subjects, which is more visible in the side view.
    Under such a standard body pose, the canonical feature then depicts the unique body shape, which is consistent within a subject but different between subjects.
    The appearance feature faithfully recovers the color and texture of clothing, at the standard body pose specified by the canonical feature.
    The pose feature captures the walking pose of the input frame.
    Finally, combining all three features can closely reconstruct the original input.
    This shows that our disentanglement not only preserves all information of the input, but also fulfills all the desired properties described in Tab.~\ref{tab:motivation}.

    \Paragraph{Visualization of Features in Two Frames}
    As shown in Fig.~\ref{fig:decodeimages_cross}, each result is generated by pairing the pose-irrelevant feature $\{\fap, \fca\}$ in the first column, and the pose feature $\fpo$ in the first row.
    The synthesized images show that indeed pose-irrelevant feature contributes all the appearance and body information, \emph{e.g.}, cloth, body width, as they are consistent across each row.
    Meanwhile, $\fpo$ contributes all the pose information, \emph{e.g.}, positions of hand and feet, which share similarity across columns.
    Despite that concatenating vectors from different subjects may create samples outside the input distribution of $\Dec$, the visual quality of synthetic images shows that $\Dec$ is versatile to these new samples.




    \begin{figure}[t!]
        \centering
        \subfloat[]{\includegraphics[trim=87 77 70 80, clip,width=4cm]{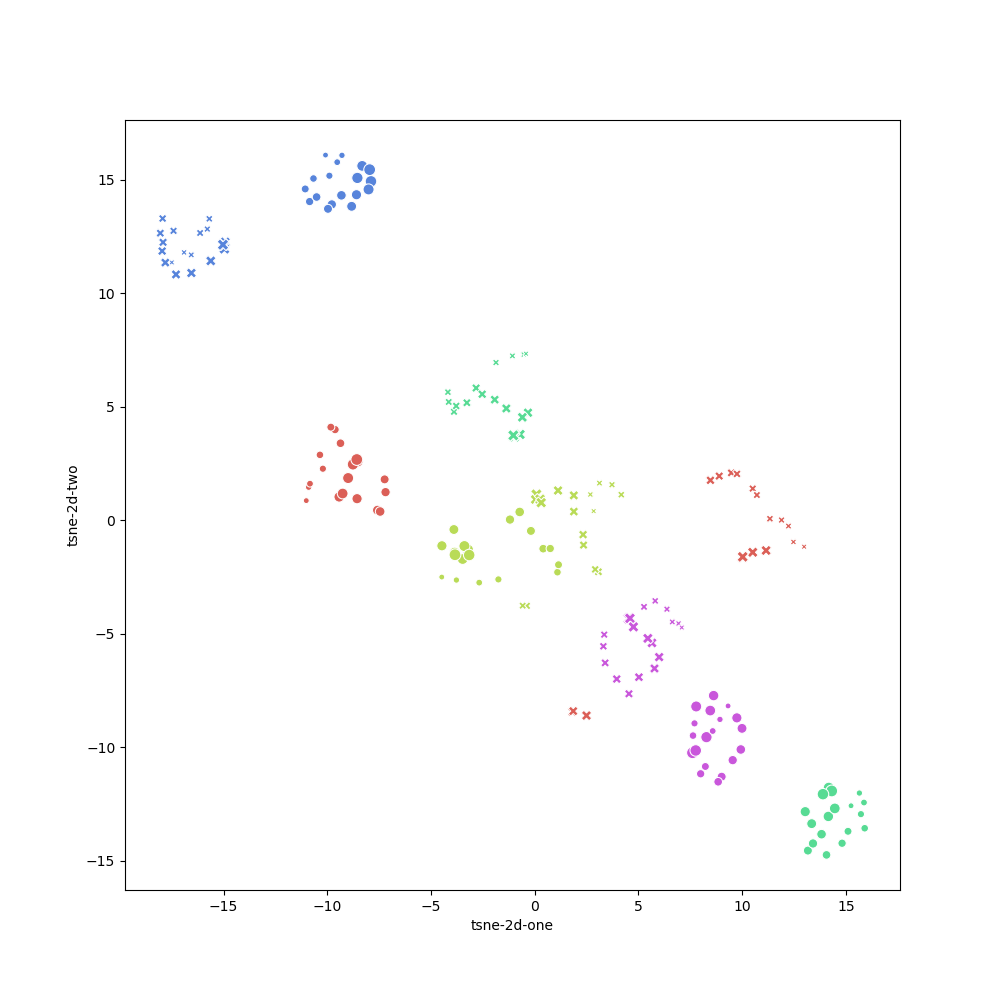}}%
        \subfloat[]{\includegraphics[trim=87 77 70 80, clip,width=4cm]{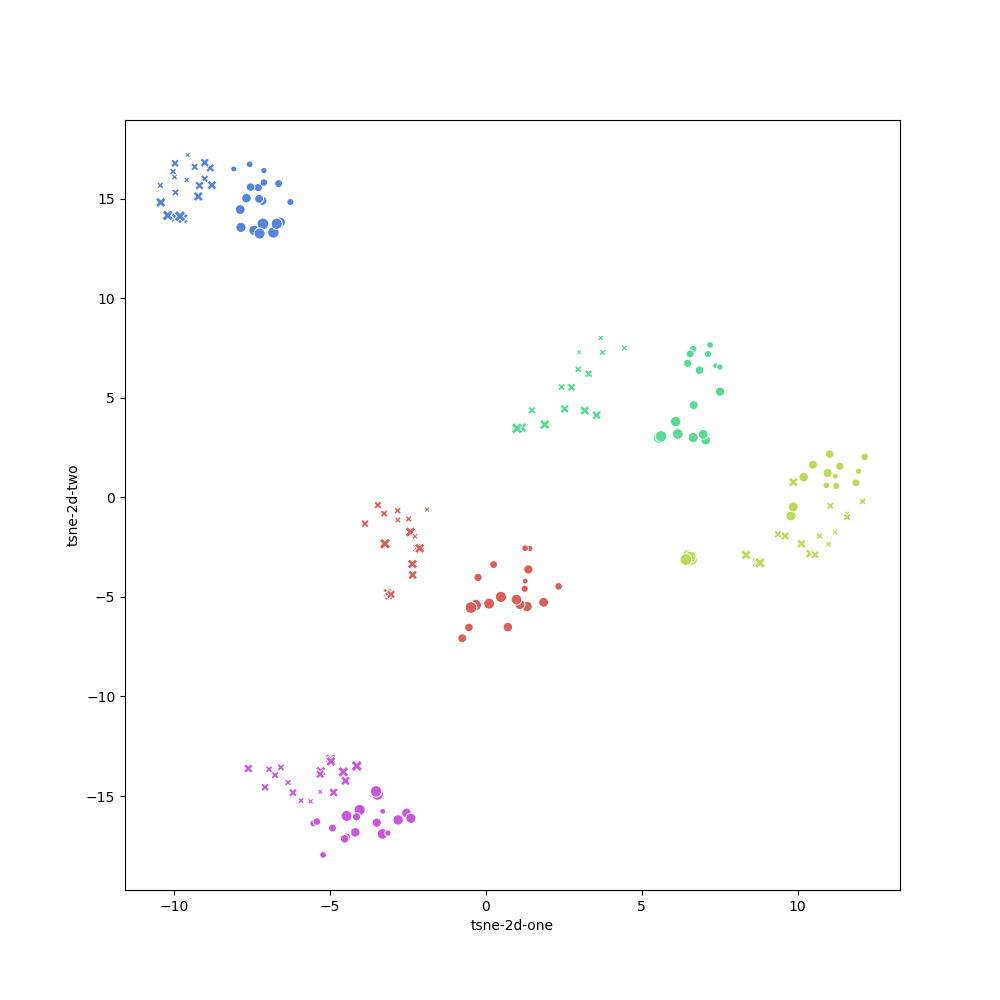}}\\ \vspace{-3mm}
        \subfloat[]{\includegraphics[trim=87 77 70 80, clip,width=4cm]{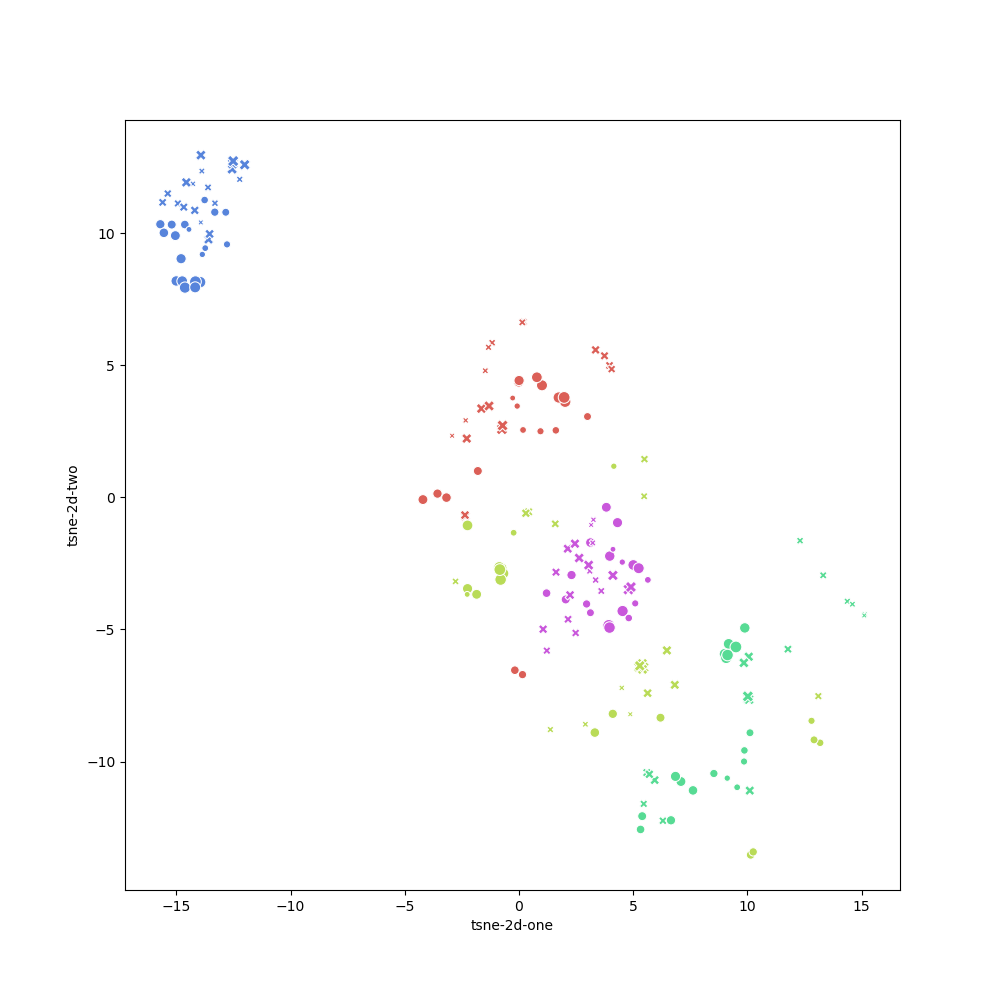}}%
        \subfloat[]{\includegraphics[trim=87 77 70 80, clip,width=4cm]{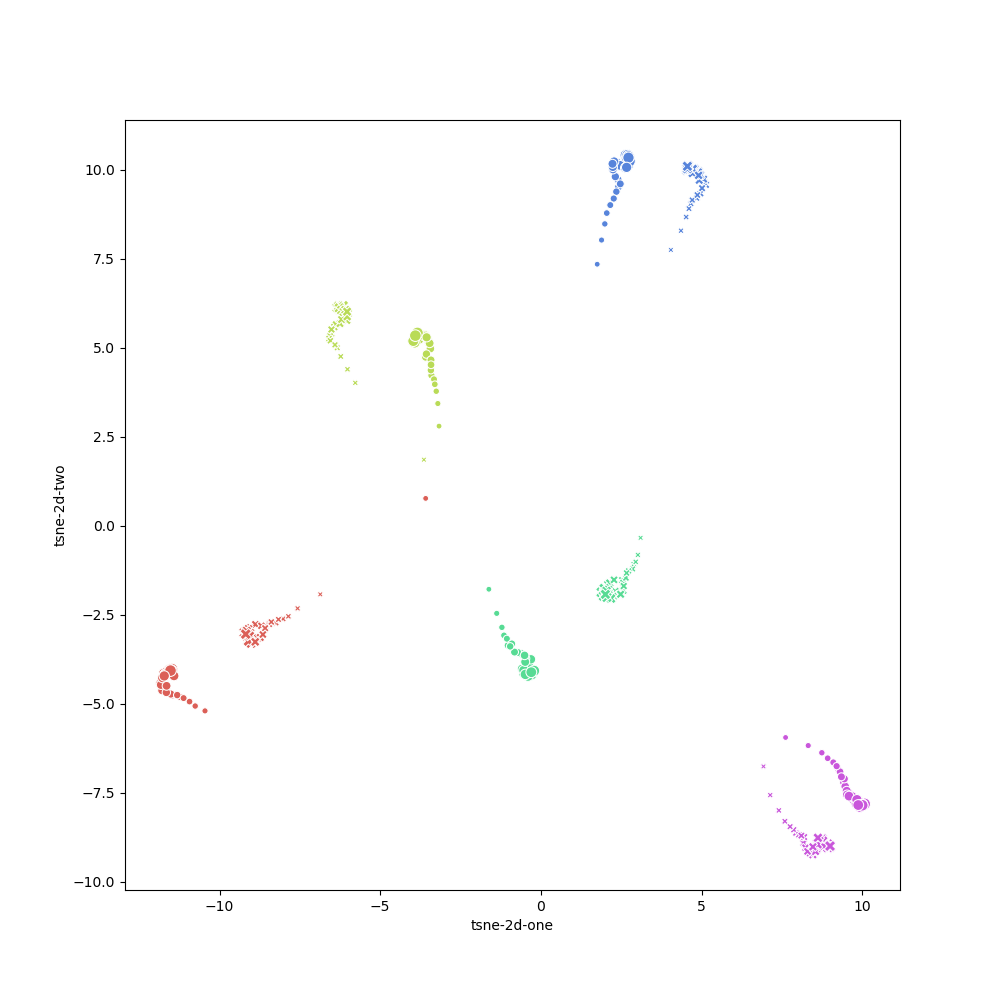}}%
        \caption{
        The t-SNE visualization of (a) appearance features $\fap$, (b) canonical features $\fca$, (c) pose features $\fpo$, and (d) dynamic gait features $\fdyngait$.
        We select $5$ subjects each with two videos of NM vs.~CL conditions.
        Each point represents a single frame, whose color is for subject ID, shape of `dot' and `cross' is NM and CL respectively, and size is frame index.
        We see that $\fca$ and $\fdyngait$ are far more discriminative than $\fap$ and $\fpo$.
        }%
        \label{fig:tsne-fafcfplstm}
    \end{figure}

    \subsubsection{Feature Visualization Through t-SNE}
    To gain more insight into the frame-level features $\fap$, $\fca$, $\fpo$ and sequence-level LSTM feature aggregation, we apply t-SNE~\cite{maaten2008visualizing} to these features to visualize their distribution in a $2$D space.
    With the learnt models in Sec.~\ref{sec:fv_syn}, we randomly select two videos under NM and CL conditions for each of $5$ subjects.

    Fig.~\ref{fig:tsne-fafcfplstm} (a,b) visualizes the $\fap$ and $\fca$ features.
    Obviously, for the appearance feature $\fap$, the margins between intra-class and inter-class distances are unpromising, which shows that $\fap$ has limited discrimination power.
    In contrast, the canonical feature $\fca$ has both the compact intra-class variations and separable inter-class differences -- useful for identity classification.
    In addition, we visualize the $\fpo$ from $\Enc$ and its corresponding $\fdyngait$ at each time step in Fig.~\ref{fig:tsne-fafcfplstm} (c-d).
    As defined in Eqn.~\ref{eq4}, we enforce the averaged $\fpo$ of the same subject to be consistent under different conditions.
    Since Eqn.~\ref{eq4} only minimizes the intra-class distance, it cannot guarantee the discrimination among subjects. However, after aggregation by the LSTM network, distances of points at longer time duration for inter-class are substantially enlarged.

    \subsubsection{Loss Function's Impact on Performance}

\begin{figure}[t!]
        \vspace{-2mm}
        \includegraphics[width=8.5cm]{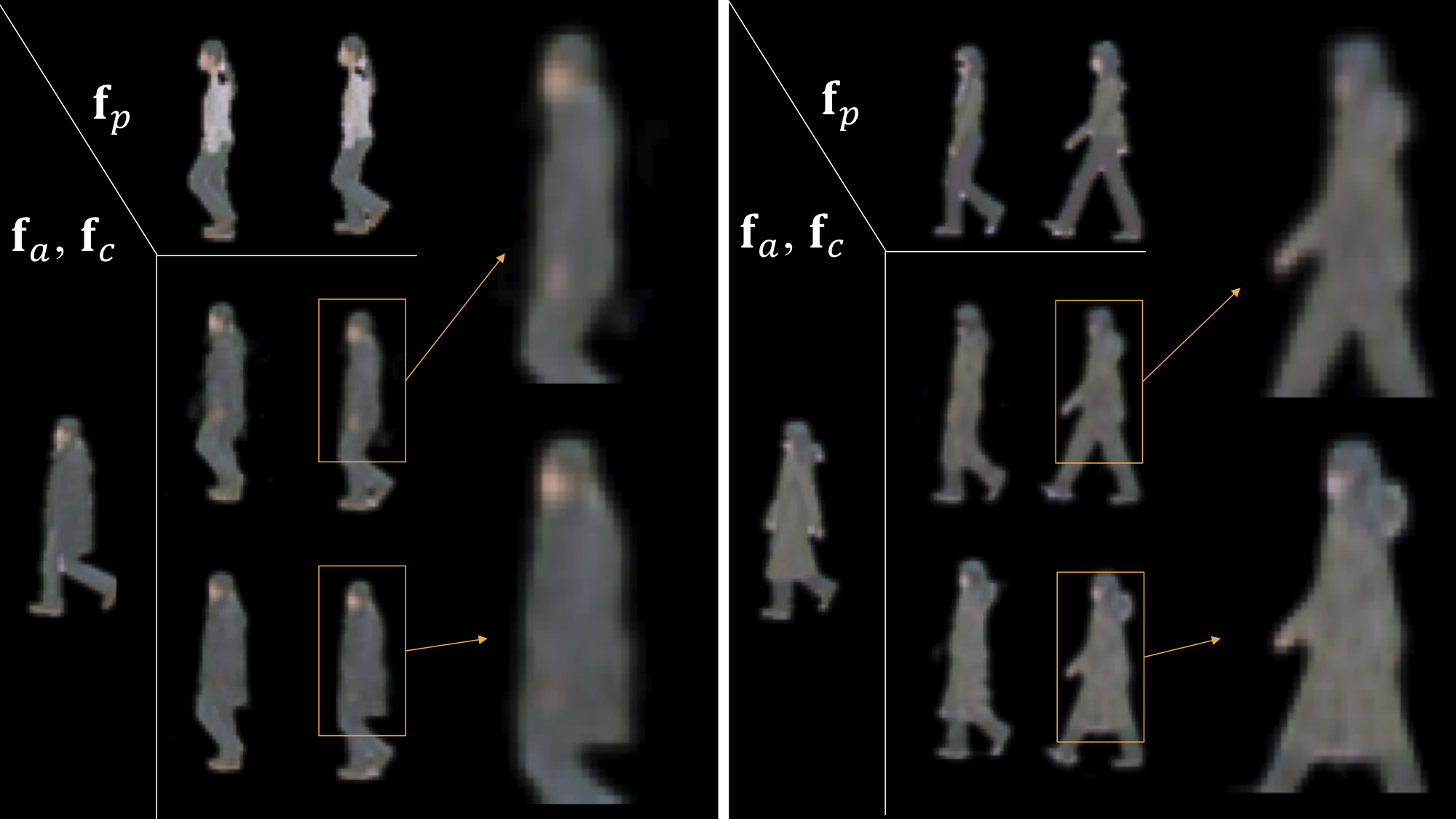}
        \centering
        \caption{
        Synthesis on CASIA-B by decoding pose-irrelevant feature $\{\fap,\fca\}$ and pose feature $\fpo$ from videos under NM vs.~CL conditions.
        Left and right parts are two examples.
        For each example, $\{\fap,\fca\}$ is extracted from the first column (CL) and $\fpo$ is from the top row (NM).
        Top row synthetic images are generated from model trained {\it without} $\L_{\text{pose-sim}}$ loss, bottom row is {\it with} the loss.
        To show the difference, details in synthetic images are magnified.
        }
        \label{fig:PoseLossCompare}
    \end{figure}

    \Paragraph{Disentanglement with Pose Similarity Loss}
    With the cross reconstruction loss, the appearance feature $\fap$ and canonical feature $\fca$ can be enforced to represent static information that shares across the video.
    However, as discussed, $\fpo$ could be contaminated by the appearance information or even encode the entire video frame.
    Here we show the benefit of the pose similarity loss $\L_{\text{pose-sim}}$ to feature disentanglement.
    Fig.~\ref{fig:PoseLossCompare} shows the cross visualization of two different models learned with and without $\L_{\text{pose-sim}}$.
    Without $\L_{\text{pose-sim}}$ the decoded image shares some appearance and body characteristic, \emph{e.g.}, cloth style, contour, with $\fpo$.
    Meanwhile, with $\L_{\text{pose-sim}}$, appearance better matches with $\fap$ and $\fca$. 


    \begin{table}[t]
        \renewcommand\arraystretch{1.2}
        \caption{Ablation study on various options of the disentanglement loss, classification loss, and classification features.
        A GaitNet model is trained on NM and CL conditions of lateral view with the first $74$ subjects of CASIA-B and tested on remaining subjects.
        }
        \label{tab:ablation}
        \centering
        \resizebox{\linewidth}{!}{
        \begin{tabular}{cccc}
            \toprule
            Disentanglement Loss & Classification Loss & Classification Feature & Rank-$1$\\ \midrule
            - & $\Lid$ & $\fdyngait$ & $56.0$ \\
            $\Lxrecon $ & $\Lid$  & $\fdyngait$ & $60.2$ \\
            $\Lxrecon + \Lpose $ & $\Lid$ & $\fdyngait$  & $85.6$\\
            \hline
            $\Lxrecon + \Lpose + \Lcano$ & $\L_{\text{id-single}}$  & $\fdyngait$ \& $\fstagait$ &$72.5$   \\ 
            $\Lxrecon + \Lpose + \Lcano$ & $\L_{\text{id-ae}}$~\cite{srivastava2015unsupervised}  & $\fdyngait$ \& $\fstagait$ & $76.5$   \\ 
            $\Lxrecon + \Lpose + \Lcano$ & $\L_{\text{id-avg}}$  & $\fdyngait$ \& $\fstagait$ & $82.6$   \\ 
            \hline
            $\Lxrecon + \Lpose + \Lcano$ & $\Lid$ & $\fap$  & $33.4$\\
            $\Lxrecon + \Lpose + \Lcano$ & $\Lid$ &$\fstagait$ & $76.3$ \\
            $\Lxrecon + \Lpose + \Lcano$ & $\Lid$ & $\fdyngait$ & $85.9$ \\

            $\Lxrecon + \Lpose + \Lcano$ & $\Lid$ & $\fdyngait$ \& $\fstagait$  & $\mathbf{92.1}$\\
            \bottomrule
        \end{tabular}}
    \end{table}


    \begin{figure}%
        \centering
        \subfloat[]{\includegraphics[trim=87 77 70 80, clip,width=3cm]{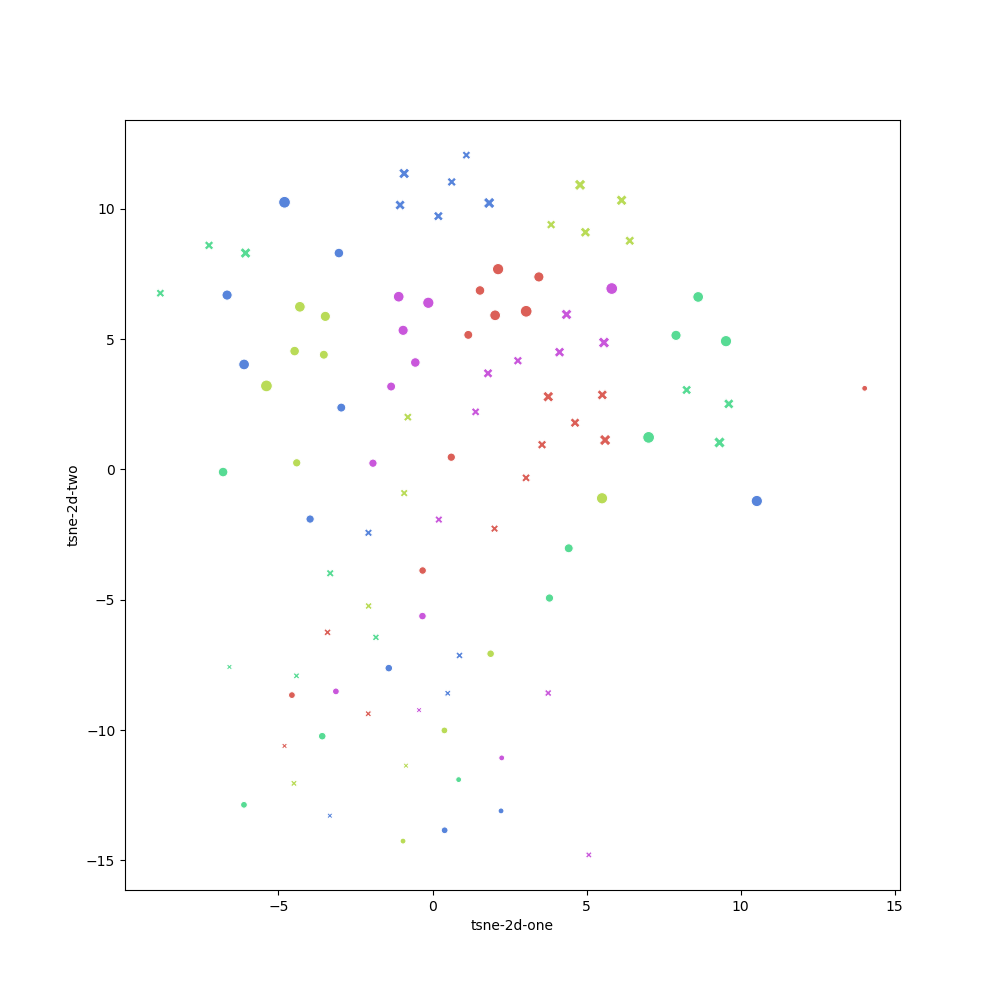}}%
        \subfloat[]{\includegraphics[trim=87 77 70 80, clip,width=3cm]{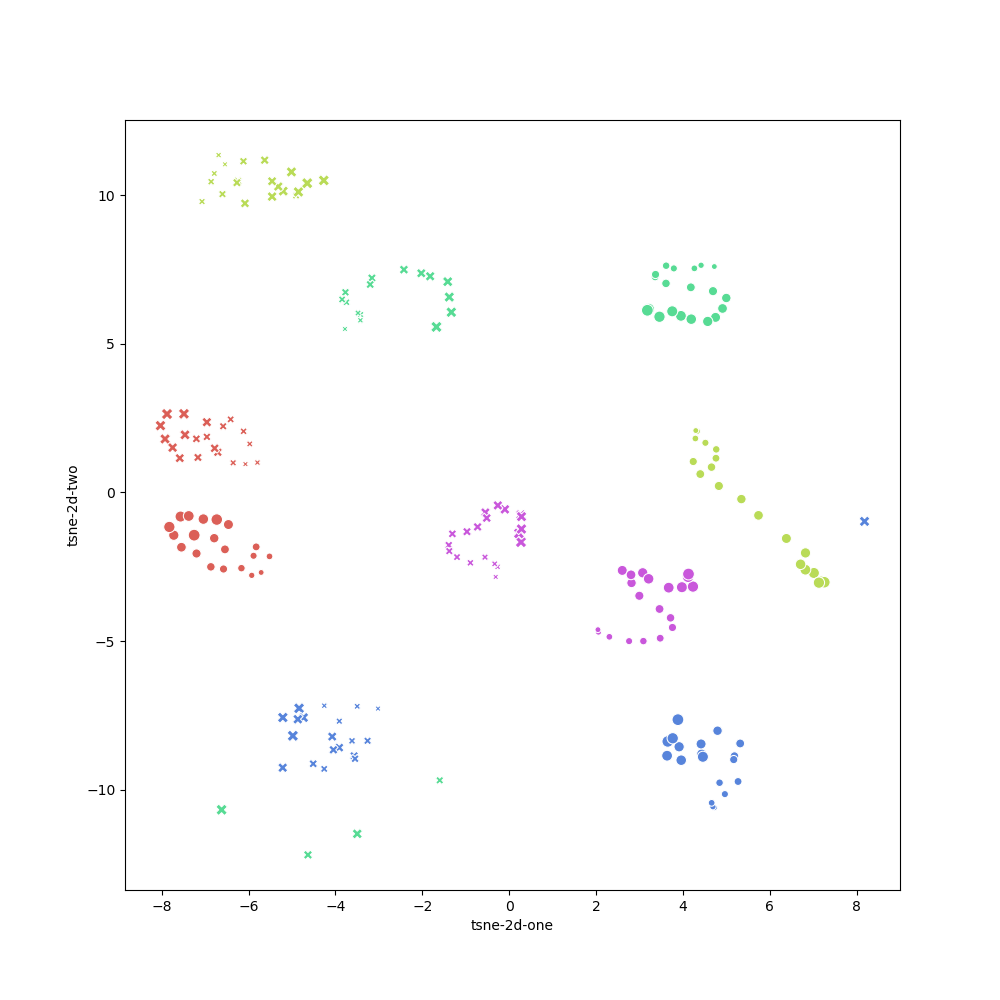}}%
        \subfloat[]{\includegraphics[trim=87 77 70 80, clip,width=3cm]{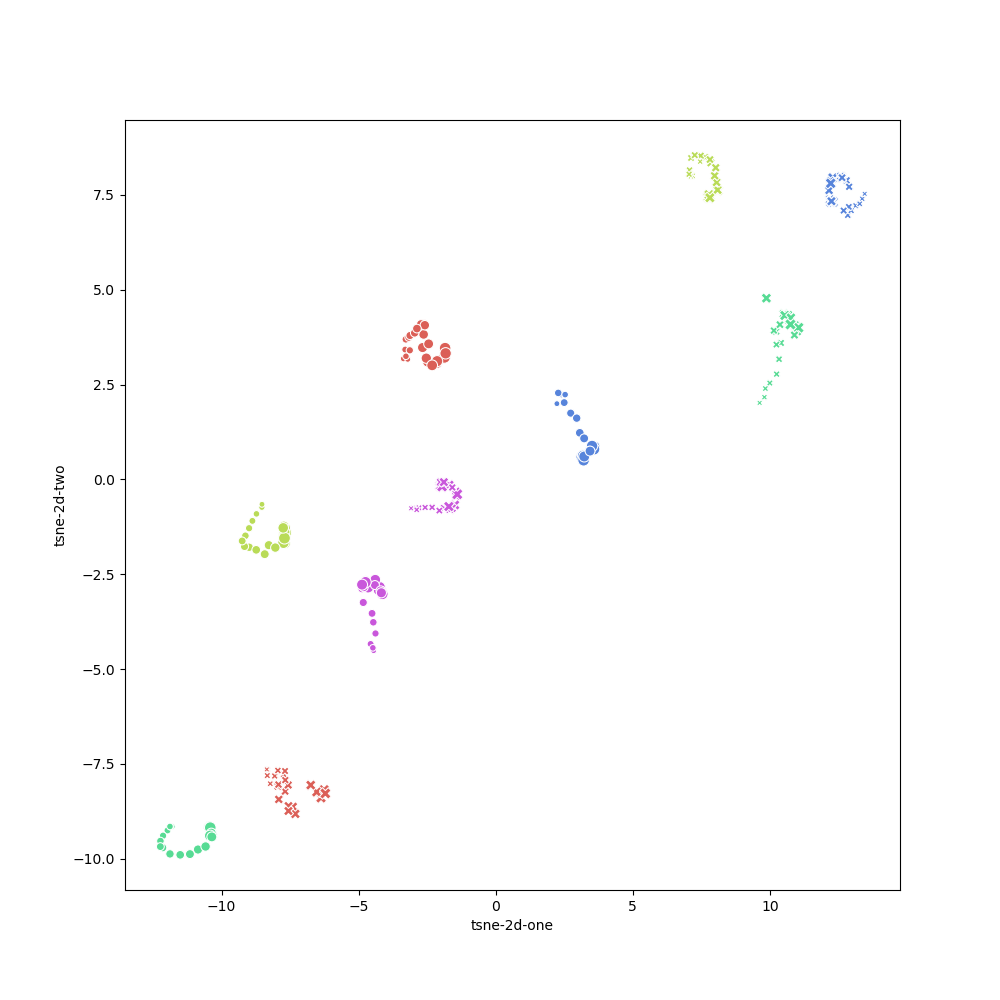}}%
        \qquad
        \subfloat[]{\includegraphics[trim=87 77 70 80, clip,width=3cm]{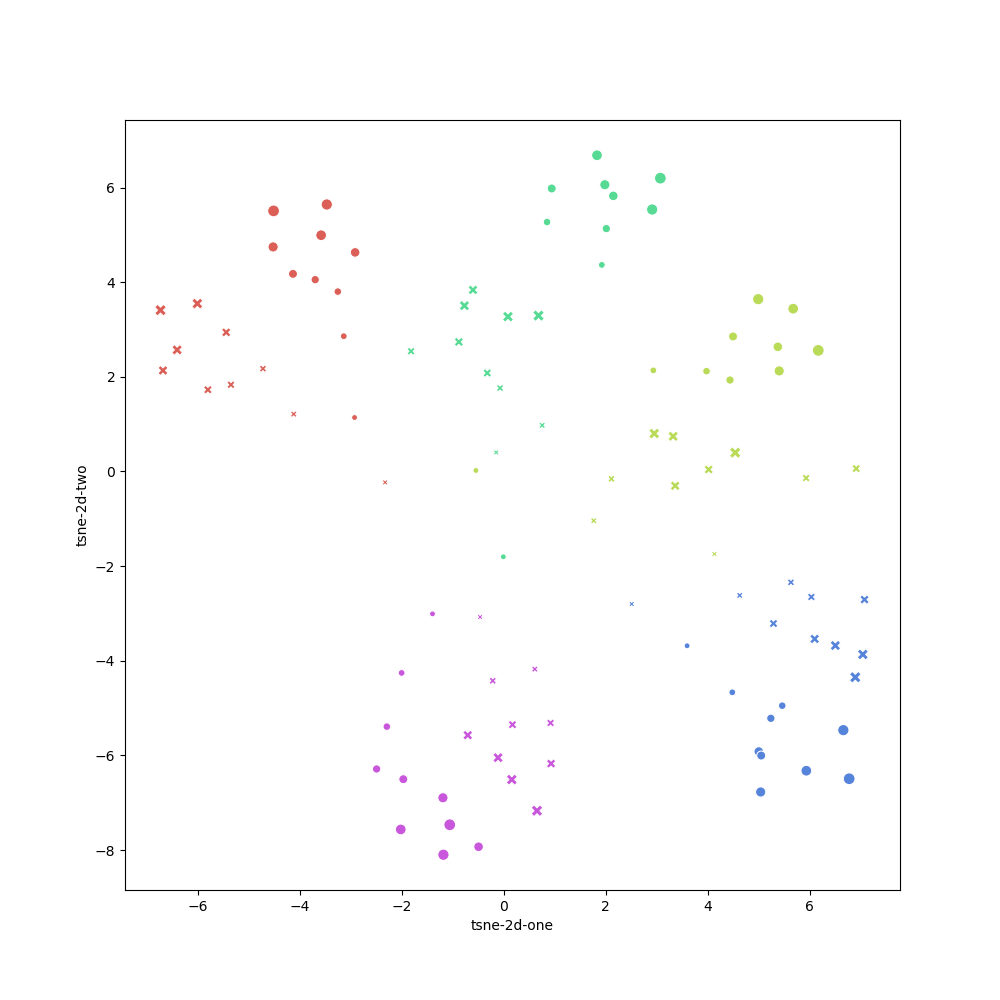}}%
        \subfloat[]{\includegraphics[trim=87 77 70 80, clip,width=3cm]{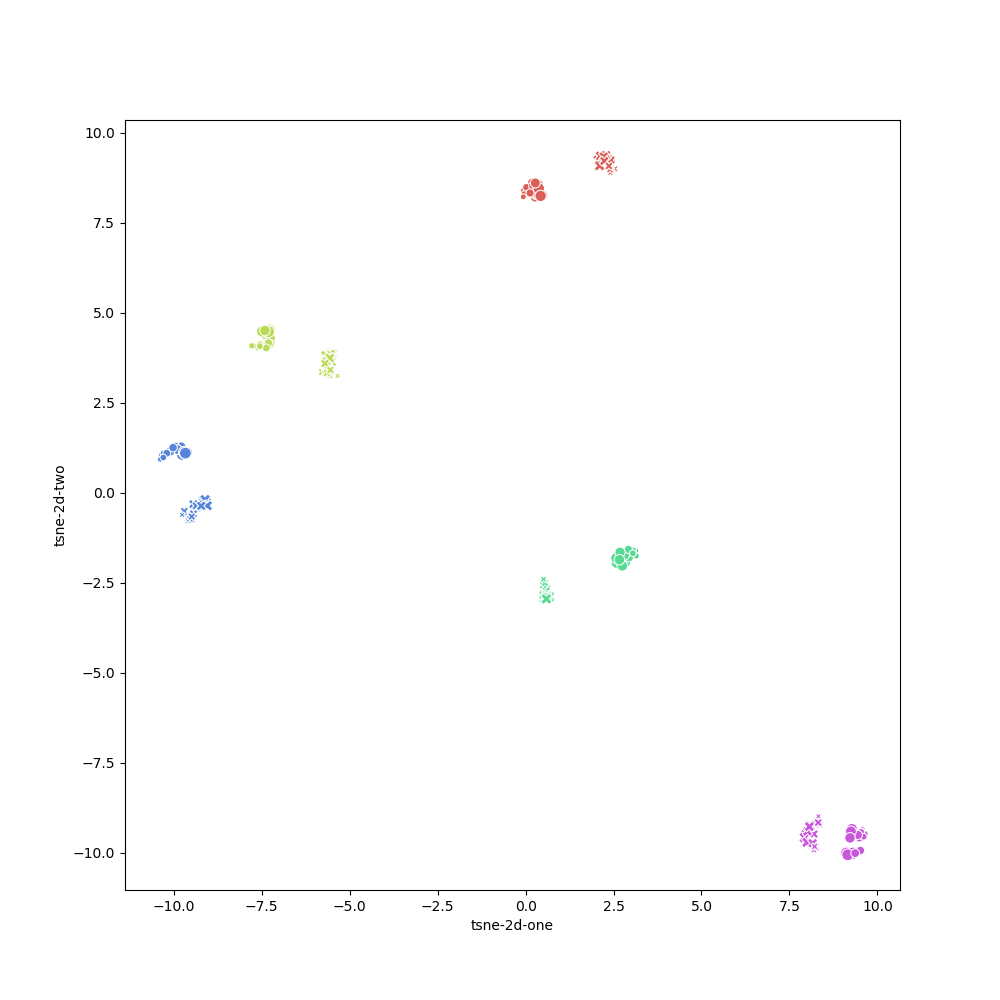}}%
        \subfloat[]{\includegraphics[trim=87 77 70 80, clip,width=3cm]{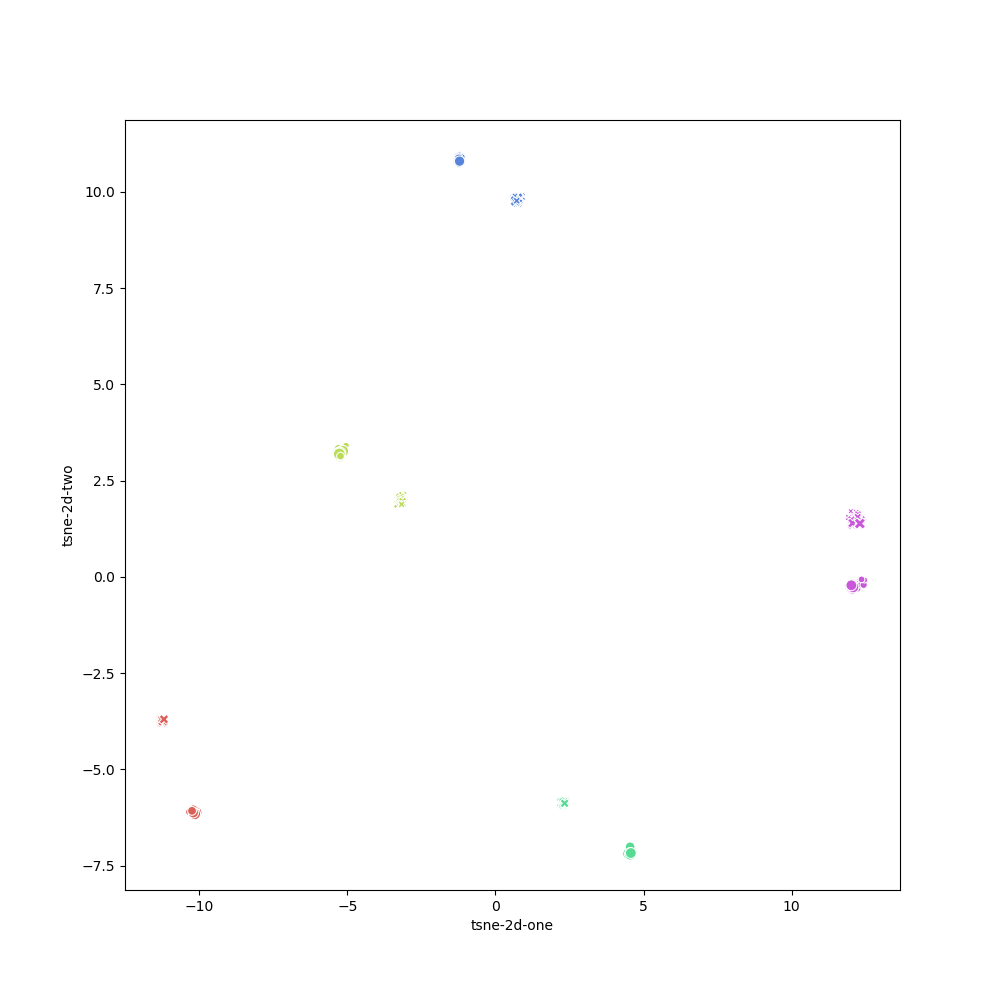}}%
        \caption{
        The t-SNE visualization of $\fdyngait$ from $5$ subjects, each with $2$ videos (NM vs.~CL).
        The symbols are defined the same as Fig.~\ref{fig:tsne-fafcfplstm}.
        The top and bottom rows are two models learnt with $L_{\text{id-single}}$ and $\Lid$ loss respectively.
        From left to tight, the points are $\fdyngait$ of the first $10$ frames, $10$-$30$ frames, and $30$-$60$ frames.
        Learning with $\Lid$ leads to more discriminative dynamic features for the entire duration.
        }%
        \label{fig:LSTM-avgvslast}
    \end{figure}

    \Paragraph{Loss Function's Impact on Recognition Performance}
    As there are various options in designing our framework, we ablate their effect on the final recognition performance from three perspectives: the disentanglement loss, the classification loss, and the classification feature.
    Tab.~\ref{tab:ablation} reports the Rank-$1$ recognition accuracy of different variants of our framework on CASIA-B under NM vs.~CL and lateral view.
    The model is trained with all videos of the first $74$ subjects and tested on the remaining $50$ subjects.

    We first explore the effects of different disentanglement losses applied to $\fdyngait$ and use $\fdyngait$ only for classification.
    Using $\Lid$ as the classification loss, we train different variants of our framework: a baseline without any disentanglement losses, a model with $\Lxrecon$ and our model with both $\Lxrecon$ and $\Lpose$.
    The baseline achieves the accuracy of $56.0\%$.
    Adding $\Lxrecon$ slightly improves the accuracy to $60.2\%$.
    By combining with $\Lpose$, our model significantly improves the accuracy to $85.6\%$.
    Between $\Lxrecon$ and $\Lpose$, the pose similarity loss plays a more critical role as $\Lxrecon$ is mainly designed to constrain the appearance feature, which does not directly benefit identification. 

    We also compare the effects of different classification losses applied to $\fdyngait$.
    Even though the classification loss only affects $\fdyngait$, we report the performance with both $\fdyngait$ and $\fstagait$ for a direct comparison with our full model in the last row.
    With the disentanglement loss of $\Lxrecon$, $\Lpose$ and $\Lcano$, we benchmark different options of the classification loss as presented in Sec.~\ref{sec:disentangle}, as well as the autoencoder loss by Srivastava \emph{et al.}~\cite{srivastava2015unsupervised}.
    The model using the conventional identity loss on the final LSTM output $\L_{\text{id-single}}$ achieves the rank-$1$ accuracy of $72.5\%$.
    Using the average output of LSTM as the identity feature, $\L_{\text{id-avg}}$ improves the accuracy to $82.6\%$.
    The autoencoder loss~\cite{srivastava2015unsupervised} achieves a good performance of $76.5\%$.
    However, it is still far from our proposed incremental identity loss $\Lid$'s performance at $92.1\%$.
    Fig.~\ref{fig:LSTM-avgvslast} further visualizes the $\fdyngait$ over time, for two models learnt with $L_{\text{id-single}}$ and $\Lid$ loss respectively.
    Clearly, even with less than $10$ frames, the model with $\Lid$ shows more discriminativeness, which also increases rapidly as time progresses.

    Finally, we compare different features in computing the final classification score.
    The performance is based on the model with full disentanglement losses and $\Lid$ as the classification loss.
    When $\fap$ is utilized in cosine distance calculation, the rank-$1$ accuracy is merely $33.4\%$, while $\fstagait$ and $\fdyngait$ achieve $76.3\%$ and $85.9\%$ respectively. The results prove the learnt $\fca$ and $\fpo$ are effective for classification while $\fap$ has limited discriminative power.
    Also, by combining both $\fstagait$ and $\fdyngait$ features, the recognition performance can be further improved to $92.1\%$.
    We believe that such performance gain is owing to the complementary discriminative information offered by $\fstagait$ w.r.t.~$\fdyngait$. 


    \begin{figure}[t]
        \includegraphics[width=4.40cm]{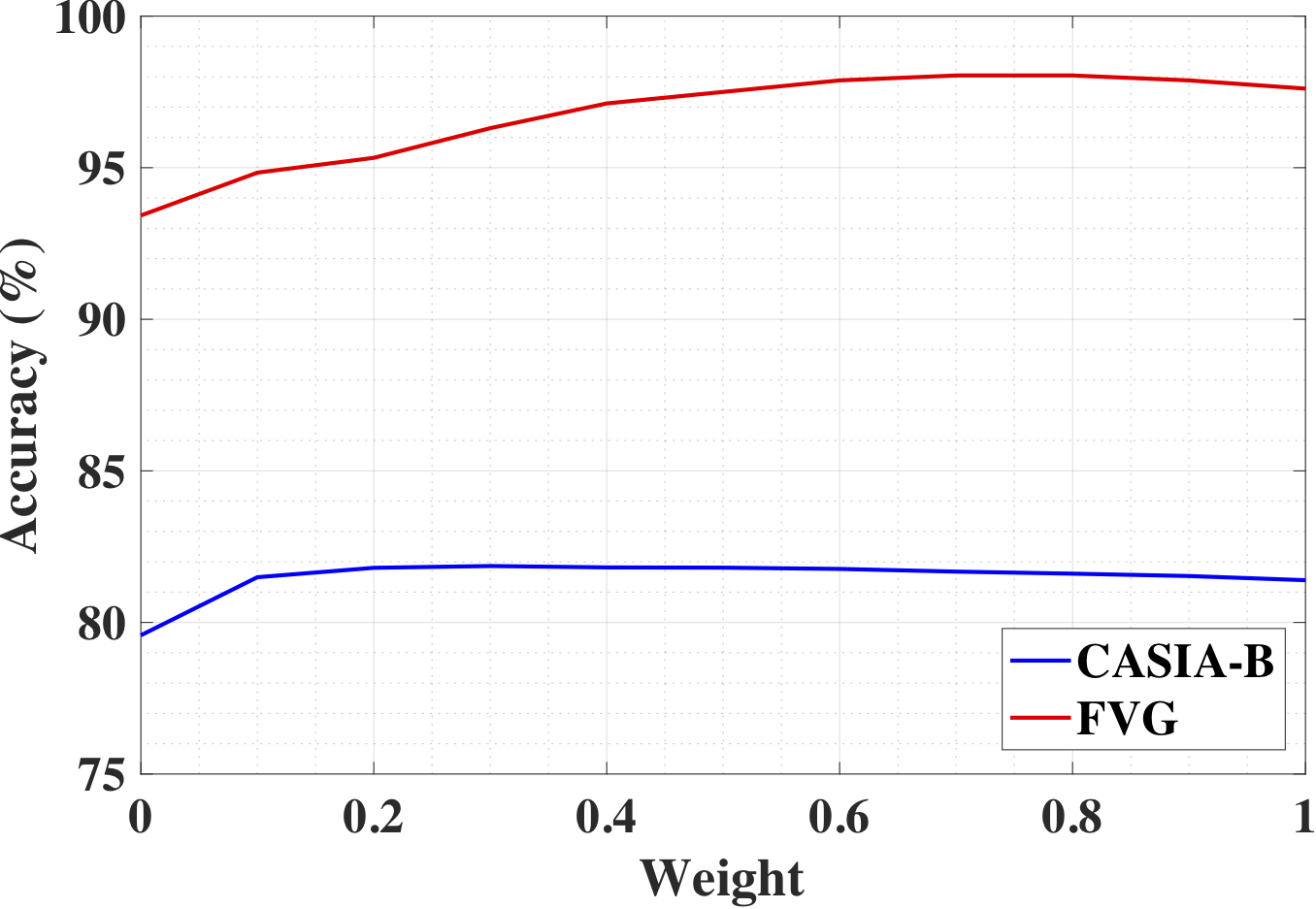}
        \centering
        \caption{ Recognition by fusing $\fdyngait$ and $\fstagait$ scores with different weights as defined in Eqn.~\ref{eq_fusion}.
        Rank-$1$ accuracy and TAR@$1\%$ FAR is calculated for CASIA-B and FVG, respectively. }
        \label{fig:alpha}
    \end{figure}


    \subsubsection{Dynamic vs.~Static Gait Features}
    \label{sec:weights}

    Since $\fdyngait$ and $\fstagait$ are complementary in classification, it is interesting to understand their relative contributions, especially in the various scenarios of gait recognition.
    This amounts to exploring a global weight $\alpha$ for the GaitNet on various training data, where $\alpha$ ranges from $0$ to $1$.
    There are three protocols on CASIA-B and hence three GaitNet models are trained respectively.
    We calculate the weighted score of all three models on the training data of protocol $1$, since it is the most comprehensive and representative protocol covering all the viewing angles and conditions.
    The same experiment is conducted on ``ALL" protocol of the FVG dataset.

    As shown in Fig.~\ref{fig:alpha}, GaitNet has the best average performance on CASIA-B when $\alpha$ is around $0.2$, while on FVG  $\alpha$ is around $0.75$.
    According to Eqn.~\ref{eq_fusion}, $\fstagait$ has relatively more classification contributions on CASIA-B.
    One potential reason is that it is more challenging to match dynamic walking poses under large range of viewing angles.
    In comparison, FVG favors $\fdyngait$.
    Since FVG is an all-frontal-walking dataset containing varying distances or resolutions, dynamic gait is relatively easier to learn with the fixed view, while $\fstagait$ might be sensitive to resolution changes.

    Nevertheless, note that in the two extreme cases, where only $\fstagait$ or $\fdyngait$ is used, there is relatively small performance gap between them. This means that either feature is effective in classification.
    Considering this observation and the balance between databases, we choose to set $\alpha{=}0.5$, which will be used in all subsequent experiments.

    \subsubsection{Gait Recognition Over Time}
    \label{sec:ablation_overtime}
\begin{figure}[t!]
        \vspace{-3mm}
        \centering
        \subfloat[]{{\includegraphics[width=4.40cm]{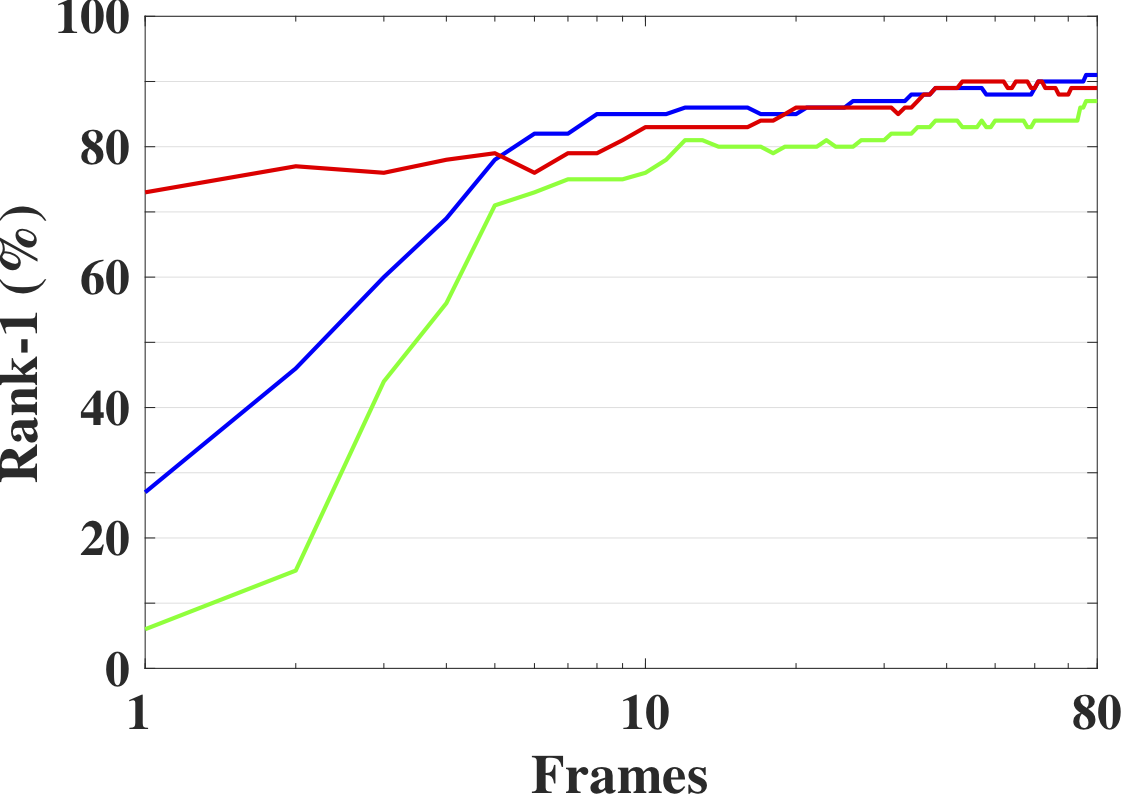} }}%
        \subfloat[]{{\includegraphics[width=4.40cm]{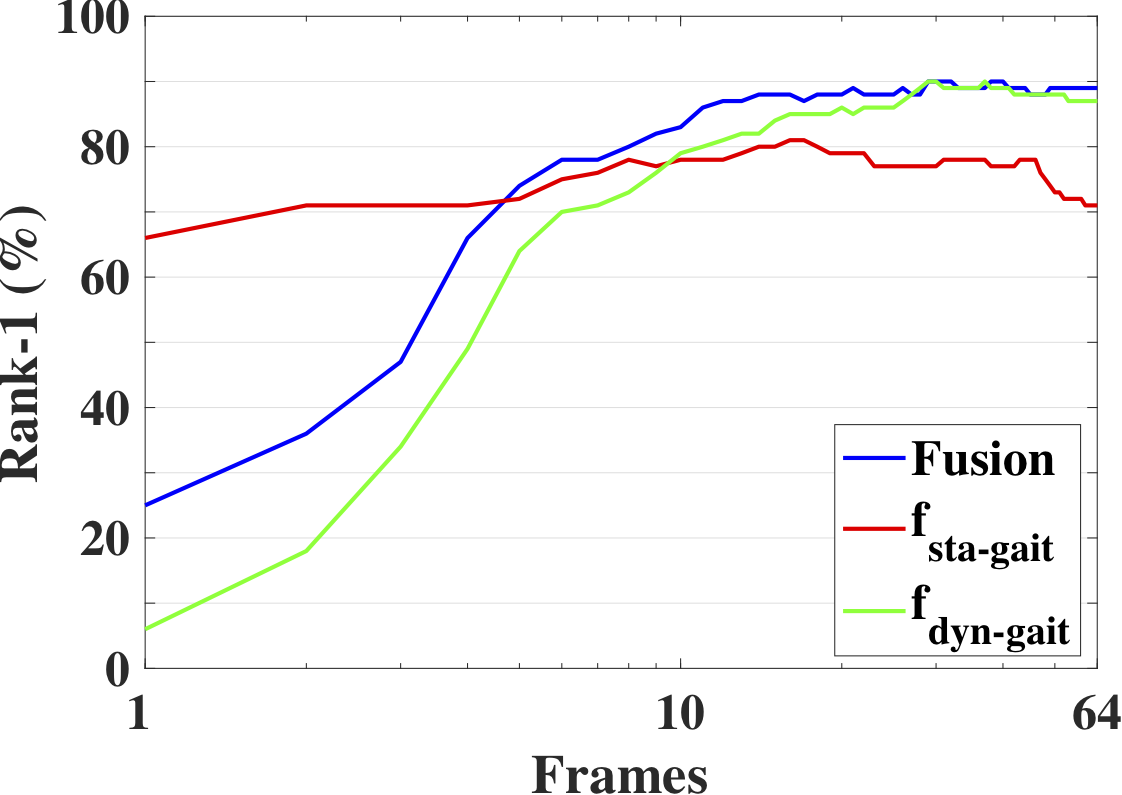} }}%
        \caption{
        Recognition performance at different video lengths. We use different feature scores ($\fstagait$, $\fdyngait$, and their fusion) on NM-{CL,BG} conditions of CAISA-B. $(a)$ is on frontal-frontal view and $(b)$ is on side-side views.
        }%
        \label{fig:dyn-sta-fusion}
    \end{figure}

   One interesting question to study is that, how many video frames are needed to achieve reliable gait recognition.
    To answer this question, we compare the performance with different feature scores ($\fstagait$, $\fdyngait$ and their fusion) for identification, with different video lengths.
    As shown in Fig.~\ref{fig:dyn-sta-fusion}, both dynamic and static features achieve stable performance starting from about $10$ frames, after which the gain in performance is relatively small.
    At $15$ FPS, a clip of $10$ frames is equivalent to merely $0.7$ seconds of walking.
    Further, the static gait feature has notable good performance even with a single video frame.
    This impressive result shows the strength of our GaitNet in processing very short clips.
    Finally, for most of the frames in this duration, the fusion outperforms both the static and dynamic gait feature alone.

    \begin{table*}[t]
        \centering
        \caption{Comparison on CASIA-B with cross view and conditions. Three models are trained for NM-NM, NM-BG, NM-CL. Average accuracies are calculated excluding probe viewing angles.}
        \label{tab:casiab-proto1}
        \resizebox{0.8\textwidth}{!}{%
        \begin{tabular}{l|ccccccccccccccc}
            \toprule
            Gallery NM \#$1$-$4$ & \multicolumn{12}{c}{$0^\circ$-$180^\circ$ (exclude identical viewing angle)} \\
            \midrule
            Probe NM \#$5$-$6$ & $0^\circ$ & $18^\circ$  & $36^\circ$ & $54^\circ$ & $72^\circ$ & $90^\circ$ & $108^\circ$ & $126^\circ$  & $144^\circ$ & $162^\circ$ & $180^\circ$  & Mean \\ \midrule

            ViDP~\cite{hu2013view} & - & - & - & $64.2$ & - & $60.4$ & - & $65.0$ & - & - & - & -  \\ 

            LB~\cite{wu2017comprehensive} & $82.6$ & $90.3$ & $96.1$ & $94.3$ & $90.1$ & $87.4$ & $89.9$ & $94.0$ & $94.7$ & $91.3$ & $78.5$ & $89.9$ \\ 

            $3$D MT network ~\cite{wu2017comprehensive} & $87.1$ & $93.2$ & $97.0$ & $94.6$ & $90.2$ & $88.3$ & $91.1$ & $93.8$ & $96.5$ & $96.0$ & $85.7$ & $92.1$ \\ 

            J-CNN~\cite{zhang2019comprehensive} & $87.2$ & $93.2$ & $96.3$ & $95.9$ & $91.6$ & $86.5$ & $89.8$ & $93.8$ & $95.1$ & $93.0$ & $80.8$ & $91.2$ \\

            GaitNet-pre~\cite{gaitnet_cvpr}& $91.2$ & 92.0 & 90.5 & $95.6$ & 86.9 &  $92.6$ & 93.5 & $96.0$ & 90.9 & $88.8$ & $89$ & $91.6$ \\

            GaitNet & $\mathbf{93.1}$ & $\mathbf{92.6}$ & $\mathbf{90.8}$ & $\mathbf{92.4}$ & $\mathbf{87.6}$ & $\mathbf{95.1}$ & $\mathbf{94.2}$ & $\mathbf{95.8}$ & $\mathbf{92.6}$ & $\mathbf{90.4}$ & $\mathbf{90.2}$   & $\mathbf{92.3}$ \\

            \bottomrule

            Probe BG \#$1$-$2$ & $0^\circ$ & $18^\circ$  & $36^\circ$ & $54^\circ$ & $72^\circ$ & $90^\circ$ & $108^\circ$ & $126^\circ$   & $144^\circ$ & $162^\circ$ & $180^\circ$ & Mean  \\ \midrule

            LB-subGEI~\cite{wu2017comprehensive} & $64.2$ & $80.6$ & $82.7$ & $76.9$ & $64.8$ & $63.1$ & $68.0$ & $76.9$ & $82.2$ & $75.4$ & $61.3$ & $72.4$ \\ 

            J-CNN~\cite{zhang2019comprehensive} & $73.1$ & $78.1$ & $83.1$ & $81.6$ & $71.6$ & $65.5$ & $71.0$ & $80.7$ & $79.1$ & $78.6$ & $68.0$ & $75.0$ \\

            GaitNet-pre~\cite{gaitnet_cvpr} & $83.0$ & 87.8 & 88.3 & $93.3$ & 82.6 &  $74.8$ & 89.5 & $91.0$  & 86.1 & $81.2$ & $85.6$ & $85.7$ \\

            GaitNet & $\mathbf{88.8}$ & $\mathbf{88.7}$ & $\mathbf{88.7}$ & $\mathbf{94.3}$ & $\mathbf{85.4}$ & $\mathbf{92.7}$ & $\mathbf{91.1}$ & $\mathbf{92.6}$ & $\mathbf{84.9}$ & $\mathbf{84.4}$ & $\mathbf{86.7}$   & $\mathbf{88.9}$ \\

            \bottomrule


            Probe CL \#1-2 & $0^\circ$ & $18^\circ$  & $36^\circ$ & $54^\circ$ & $72^\circ$ & $90^\circ$ & $108^\circ$ & $126^\circ$  & $144^\circ$ & $162^\circ$ & $180^\circ$ & Mean  \\
            \midrule
            LB-subGEI~\cite{wu2017comprehensive} & $37.7$ & $57.2$ & $66.6$ & $61.1$ & $55.2$ & $54.6$  & $55.2$ & $59.1$ & $58.9$ & $48.8$ & $39.4$ & $53.98$ \\

            J-CNN~\cite{zhang2019comprehensive}   & $46.1$ &58.4 &64.4& $64.2$ & 55.5 & $50.5$ & 54.7 & $55.8$ & 53.3 & 51.3 & 39.9 & 54.01
            \\

            GaitNet-pre~\cite{gaitnet_cvpr}   & $42.1$ & $58.2$ & $65.1$ & $70.7$ &  $68.0$  & $70.6$ & $65.3$ & $69.4$ & $51.5$  & $50.1$ & $36.6$ & $58.9$
            \\
            GaitNet & $\mathbf{50.1}$ & $\mathbf{60.7}$ & $\mathbf{72.4}$ &$\mathbf{72.1}$ & $\mathbf{74.6}$ & $\mathbf{78.4}$ & $\mathbf{70.3}$ & $\mathbf{68.2}$ & $\mathbf{53.5}$  & $\mathbf{44.1}$ & $\mathbf{40.8}$ &$\mathbf{62.3}$
            \\
            \bottomrule

        \end{tabular}%
        }
    \end{table*}


    \begin{table*}[t]
        \centering
        \caption{Recognition accuracy cross views under NM on CASIA-B dataset. One single GaitNet model is trained for all the viewing angles. }
        \label{tab:casiab-proto2}
        \resizebox{0.8\textwidth}{!}{
        \begin{tabular}{lccccccccccc}
            \toprule
            Methods & $0^\circ$  & $18^\circ$ & $36^\circ$        & $54^\circ$            & $72^\circ$   & $108^\circ$  & $126^\circ$ & $144^\circ$ & $162^\circ$ & $180^\circ$ & Average \\ \midrule
            CPM~\cite{chen2018multi}                     & $13$ & $14$ & $17$        & $27$            & $62$  & $65$   & $22$  & $20$  & $15$  & $10$  & $24.1$   \\ 
            GEI-SVR~\cite{kusakunniran2010support}         & $16$ & $22$ & $35$        & $63$            & $95$   & $95$   & $65$  & $38$  & $20$  & $13$  & $42.0$      \\ 
            CMCC~\cite{kusakunniran2014spatiot}            & $18$ & $24$ & $41$        & $66$            & $96$   & $95$   & $68$  & $41$  & $21$  & $13$  & $43.9$   \\ 
            ViDP~\cite{hu2013view}                       & $8$ & $12$ & $45$        & $80$            & $\mathbf{100}$  & $100$  & $81$  & $50$  & $15$  & $8$   & $45.4$   \\ 
            STIP+NN~\cite{kusakunniran2014recognizing}         & - & - & - & - & $84.0$   & $86.4$ & - & - & - & - & -    \\ 
            LB~\cite{wu2017comprehensive}    & $18$ & $36$ & $67.5$      & $93$            & $99.5$ & $99.5$  & $92$  & $66$  & $36$  & $18$  & $56.9$   \\ 
            L-CRF~\cite{chen2018multi}                   & $38$ & $75$ & $68$        & $93$            & $98$   & $99$   & $93$  & $67$  & $76$  & $39$  & $67.8$   \\ 
            GaitNet-pre~\cite{gaitnet_cvpr}                                    & $68$   &74 & $88$ & $91$ & $99$     &$98$      &$84$     &$75$     &$76$     &$65$     & $81.8$        \\
            GaitNet & $\mathbf{82}$   &$\mathbf{83}$ & $86$ &$91$  & $93$     &$98$      &$\mathbf{92}$     &$\mathbf{90}$     &$\mathbf{79}$     &$\mathbf{79}$     & $\mathbf{87.3}$        \\


            \bottomrule
        \end{tabular}}
    \end{table*}

    \subsection{Evaluation on Benchmark Datasets}
    \subsubsection{CASIA-B}
    \label{sec:CASIA-B}

    Since various experimental protocols have been defined on CASIA-B, for a fair comparison, we strictly follow the respective protocols in the baseline methods.
    Following~\cite{wu2017comprehensive}, Protocol $1$ uses the first $74$ subjects for training and remaining $50$ for testing, regarding variations of NM (normal), BG (carrying bag) and CL (wearing a coat) with crossing viewing angles of $0^\circ$ to $180^\circ$. Three models are trained for comparison in Tab.~\ref{tab:casiab-proto1}.
    For the detailed protocol, please refer to~\cite{wu2017comprehensive}.
    Here we mainly compare our work to Wu \emph{et al.}~\cite{wu2017comprehensive}, along with other methods~\cite{hu2013view,zhang2019comprehensive}.
    We denote our preliminary work~\cite{gaitnet_cvpr} as GaitNet-pre and this work as GaitNet.
    Under multiple viewing angles and across three variations, GaitNet achieves the best performance compared to all SOTA methods and GaitNet-pre since $\fca$ can distill more discriminative information under various viewing angles and conditions.


    Recently, Chen \emph{et al.}~\cite{wu2017comprehensive} propose new protocols to unify the training and testing where only one single model is trained for each protocol.
    Protocol $2$ focuses on walking direction variations, where all videos used are in the NM subset.
    The training set includes videos of first $24$ subjects in all viewing angles.
    The rest $100$ subjects are for testing.
    The gallery is made of four videos at $90^\circ$ view for each subject.
    The first two videos from remaining viewing angles are the probe.
    The Rank-$1$ recognition accuracies are reported in Tab.~\ref{tab:casiab-proto2}.
    GaitNet achieves the best average accuracy of $87.3\%$ across $10$ viewing angles, with significant improvement on extreme views compared to our preliminary work~\cite{gaitnet_cvpr}.
    For example, at viewing angles of $0^\circ$, and $180^\circ$, the improvement margins are both $14\%$.
    This shows that more discriminative gait information, such as a canonical body shape information, under different views are learned in $\fca$, which contributes to the final recognition accuracy. 


    \begin{table}[t!]
        \centering
        \caption{Comparison with~\cite{chen2018multi} and~\cite{wu2017comprehensive} under different walking conditions on CASIA-B by accuracies. One single GaitNet model is trained with all gallery and probe views and the two conditions. }
        \label{tab:casiab-proto3}
        \resizebox{0.5\textwidth}{!}
        {%
        \begin{tabular}{cccccccc}
            \hline
            Probe & Gallery & \makecell{GaitNet} & \makecell{GaitNet-pre \\\cite{gaitnet_cvpr}} & \makecell{JUCNet\\ \cite{Zhang_2019_CVPR}} & \makecell{L-CRF\\ \cite{chen2018multi}} & \makecell{LB\\ \cite{wu2017comprehensive}} & \makecell{RLTDA \\ \cite{hu2013enhanced}}
            \\ \toprule
            \multicolumn{2}{c}{Subset} & \multicolumn{6}{c}{BG} \\ \hline
            $54$ & $36$ & $93.5$ & $91.6$ & $91.8$ & $\mathbf{93.8}$ & $92.7$ & $80.8$ \\ \cline{1-2}
            $54$ & $72$ & $\mathbf{94.1}$ & $90.0$ & $93.9$ & $91.2$ & $90.4$ & $71.5$ \\ \cline{1-2}
            $90$ & $72$ & $\mathbf{98.6}$ & $95.6$ & $95.9$ & $94.4$ & $93.3$ & $75.3$ \\ \cline{1-2}
            $90$ & $108$ & $\mathbf{99.3}$ & $87.4$ & $95.9$ & $89.2$ & $88.9$ & $76.5$ \\ \cline{1-2}
            $126$ & $108$ & $\mathbf{99.5}$ & $90.1$ & $93.9$ & $92.5$ & $93.3$ & $66.5$ \\ \cline{1-2}
            $126$ & $144$ & $\mathbf{90.0}$ & $93.8$ & $87.8$ & $88.1$ & $86.0$ & $72.3$ \\ \hline
            \multicolumn{2}{c}{Mean} & $\mathbf{95.8}$ & $91.4$ & $93.2$ & $91.5$ & $90.8$ & $73.8$ \\ \bottomrule
            \\
            \toprule
            \multicolumn{2}{c}{Subset} & \multicolumn{6}{c}{CL} \\ \hline
            $54$ & $36$ & $\mathbf{97.5}$ & $87.0$ & - & $59.8$ & $49.7$ & $69.4$ \\ \cline{1-2}
            $54$ & $72$ & $\mathbf{98.6}$ & $90.0$ & - & $72.5$ & $62.0$ & $57.8$ \\\cline{1-2}
            $90$ & $72$ & $\mathbf{99.3}$ & $94.2$ & - & $88.5$ & $78.3$ & $63.2$ \\\cline{1-2}
            $90$ & $108$ & $\mathbf{99.6}$ & $86.5$ & - & $85.7$ & $75.6$ & $72.1$ \\\cline{1-2}
            $126$ & $108$ & $\mathbf{98.3}$ & $89.8$ & - & $68.8$ & $58.1$ & $64.6$ \\\cline{1-2}
            $126$ & $144$ & $86.6$ & $\mathbf{91.2}$ & - & $62.5$ & $51.4$ & $64.2$ \\\cline{1-2}
            \multicolumn{2}{c}{Mean} & $\mathbf{96.7}$ & $89.8$ & - & $73.0$ & $62.5$ & $65.2$ \\ \bottomrule
        \end{tabular}
        }
    \end{table}

    Protocol $3$ focuses on appearance variations.
    Training sets have videos under BG and CL. There are $34$ subjects in total with  $54^\circ$ to $144^\circ$ viewing angles.
    Different test sets are made with the different combination of viewing angles of the gallery and probe as well as the appearance condition (BG or CL).
    The results are presented in Tab.~\ref{tab:casiab-proto3}.
    Our preliminary work has comparable performance as the SOTA method L-CRF~\cite{chen2018multi} on BG subset while significantly outperforming on CL subset.
    The proposed GaitNet outperforms on both subsets.
    Note that due to the challenge of CL protocol, there is a significant performance gap between BG and CL for all methods except ours, which is yet another evidence that our gait feature has strong invariance to all major gait variations.

    Across all evaluation protocols, GaitNet consistently outperforms the state of the art. This shows the superior of GaitNet on learning a robust representation under different variations.
    It is contributed to our ability to disentangle pose/gait information from appearance variations.
    Comparing with our preliminary work, the canonical feature $\fca$ contains discriminative power which can further improve the recognition performance.


    \begin{table*}[t]
        \centering
        \caption{Definition of FVG protocols and performance comparison.
        Under each of the $5$ protocols, the first/second columns indicate the indexes of videos used in gallery/probe.}
        \label{tab:fvg}
        \resizebox{0.7\textwidth}{!}{%
        \begin{tabular}{l|cc|cc|cc|cc|cc}
            \toprule

            Protocol & \multicolumn{2}{c|}{WS} & \multicolumn{2}{c|}{BGHT} & \multicolumn{2}{c|}{CL} & \multicolumn{2}{c|}{MP} & \multicolumn{2}{c}{ALL} \\ \midrule
            \multicolumn{11}{c}{Index of Gallery \& Probe videos} \\ \midrule
            Session $1$ &$2$ & $4$-$9$    & $2$ & $10$-$12$  & - & - & - & - & $2$ & $1$,$3$-$12$ \\
            Session $2$ &$2$ & $4$-$6$          & - & - & $2$ & $7$-$9$     & $2$ & $10$-$12$                 & $2$ & $1$,$3$-$12$ \\
            Session $3$ & - & - & - & - & - & - & - & - & - & $1-12$\\ \midrule
            TAR@FAR & $1\%$       & $5\%$      & $1\% $       & $5\%$        &$1\%$       & $5\%$      & $1\%$          & $5\%$        & $1\%$       & $5\%$     \\ \midrule
            PE-LSTM &      $79.3$        &       $87.3$      &       $59.1$        &       $78.6$        &        $55.4$      &        $67.5$     &        $61.6$         &        $72.2$         & $65.4$        &      $74.1$        \\
            GEI~\cite{han2006individual}  &    $9.4$     &      $19.5$        &       $6.1$      &       $12.5$        &         $5.7$      &       $13.2$       &      $6.3$       &          $16.7$  &     $5.8$       &       $16.1$   \\
            GEINet~\cite{shiraga2016geinet} & $15.5$     &      $35.2$        &       $11.8$      &     $24.7$          &         $6.5$      &     $16.7$         &      $17.3$       &        $35.2$  &       $13.0$       &       $29.2$       \\
            DCNN~\cite{alotaibi2017improved} &  $11.0$      &       $23.6$       &     $5.7$       &       $12.7$        &       $7.0$        &       $15.9$    &      $8.1$           &        $20.9$ &       $7.9$       &     $19.0$         \\
            LB~\cite{wu2017comprehensive}   &        $53.4$      &      $73.1$       &      $23.1$         &        $50.3$       &      $23.2$        &     $38.5$        &       $56.1$          &        $74.3$ &      $40.7$        &       $61.6$       \\
            GaitNet-pre~\cite{gaitnet_cvpr} &  $91.8$ &  $96.6$ & $74.2$ & $85.1$ & $56.8$ & $72.0$ &  $92.3$  &  $97.0$ & $81.2$ & $87.8$       \\

            GaitNet
            &  $\mathbf{96.2}$ &  $\mathbf{97.5}$
            & $\mathbf{92.3}$ & $\mathbf{96.4}$
            & $\mathbf{70.4}$ & $\mathbf{87.5}$
            &  $\mathbf{92.5}$  &  $\mathbf{96.0}$
            & $\mathbf{91.9}$ & $\mathbf{96.3}$       \\




            \bottomrule
        \end{tabular}%
        }
    \end{table*}

    \subsubsection{USF}
    The original protocol of USF~\cite{sarkar2005humanid} and the methods \cite{wang2011human,xu2007marginal,guan2014reducing,aggarwal2017covariate}
    does not define a training set, which is not applicable to our method, as well as~\cite{wu2017comprehensive}, that require data to train the models.
    Hence following the experiment setting in~\cite{wu2017comprehensive}, which randomly partitions the dataset into the non-overlapping training and test sets, each with half of the subjects.
    We test on Probe A, defined in~\cite{wu2017comprehensive}, where the probe is different from the gallery by the viewpoint. We achieve the identification accuracy of $99.7\pm 0.2\%$, which is better than $99.5\pm 0.2\%$ of our preliminary work GaitNet-pre~\cite{gaitnet_cvpr}, the reported $96.7\pm0.5\%$ of LB network~\cite{wu2017comprehensive}, and $94.7\pm2.2\%$ of multi-task GAN~\cite{he2019multi}.


    \subsubsection{FVG}
    \label{sec:fvg}

    Given that FVG is a newly collected database and no reported performance from prior work, we make the efforts to implement $4$ classic or SOTA methods on gait recognition~\cite{han2006individual,shiraga2016geinet,alotaibi2017improved,wu2017comprehensive}.
    Furthermore, given the large amount of effort in human pose estimation~\cite{feng2016learning}, aggregating joint locations over time can be a good candidate for gait features.
    Therefore we define another baseline, named PE-LSTM, using pose estimation results as the input to the same LSTM and classification loss as ours.
    Using SOTA $2$D pose estimation~\cite{fang2017rmpe}, we extract $14$ joints' locations, feed to the $3$-layer-LSTM, and train with our proposed LSTM incremental loss.
    For each of $5$ baselines and our GaitNet, one model is trained with the $136$-subject training set and tested on all $5$ protocols.

    As shown in Tab.~\ref{tab:fvg}, our method shows state-of-the-art performance compared with baselines, including the recent CNN-based methods.
    Among $5$ protocols, CL is the most challenging variation as in CASIA-B.
    Comparing with all different methods, GEI based methods suffer from frontal view due to the lack of walking information.
    Again, thanks to the discriminative canonical feature $\fca$, GaitNet achieves better recognition accuracies than GaitNet-pre.
    Also, the superior performance of our GaitNet over PE-LSTM demonstrates that our feature $\fpo$ and $\fca$ does explore more discriminate information than the joints' locations alone.

    \subsection{Comparison to Face Recognition}
    Face recognition aims to identify subjects by extracting discriminative identity features, or representation, from face images.
    Due to the vigorous development in the past few years, face recognition system is one of the most studied and deployed systems in the vision community, even superior to human on some tasks~\cite{yin2018representation}.

    However, the challenge is particularly prominent in the video surveillance scenario, where low-resolution and/or non-frontal faces are acquired at a distance.
    While gait, as a behavioral biometric compared to face, might have more advantages in those scenarios since the dynamic information can be more resistant even at a lower resolution and different viewing angles.
    Especially for GaitNet, $\fstagait$ and $\fdyngait$ can have complementary contributions in changing distances, resolutions and viewing angles.
    Therefore, to explore the advantages and disadvantages of gait recognition and face recognition in surveillance scenario, we compare our GaitNet with the most recent SOTA face recognition method, ArcFace~\cite{deng2019arcface}, on the CASIA-B and FVG databases.


    Specifically, for face recognition, we first employ SOTA face detection algorithm RetinaFace~\cite{deng2019retinaface} to detect face and ArcFace to extract features for each frame of gallery and probe videos. 
    Then the features over all frames of a video are aggregated by average pooling, an effective scheme used in prior video-based face recognition work~\cite{Gong2019MARN}.
    We measure the similarity of features by their cosine distance.
    To keep consistency with above gait recognition experiments, both face and gait report TAR at $1\%$ FAR for FVG and Rank-$1$ score for CASIA-B.
    To evaluate effects of time, we use the entire sequence as gallery and partial (\emph{e.g.}, $10$\%) sequence as probe on $10$ points on the time axis ranging from $10$\% to $100$\%.


    \begin{figure}[t]
        \centering
        ~~~~~~\subfloat{{\includegraphics[width=3.8cm]{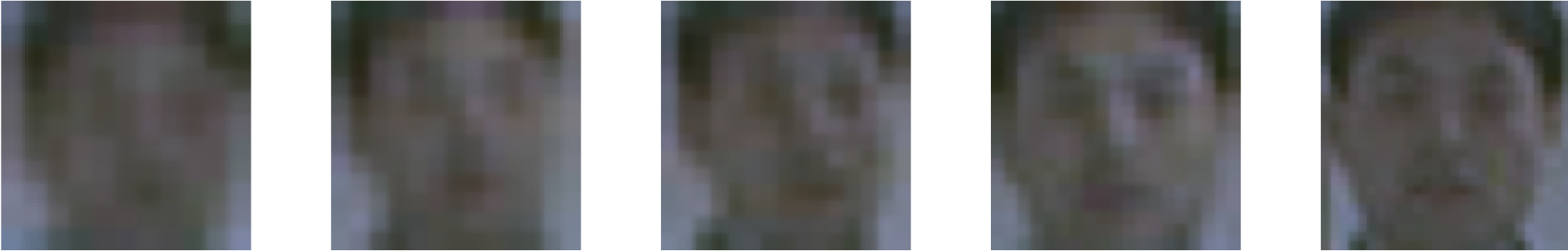} }} ~~~~~~
        \addtocounter{subfigure}{-1}
        \subfloat{{\includegraphics[width=3.8cm]{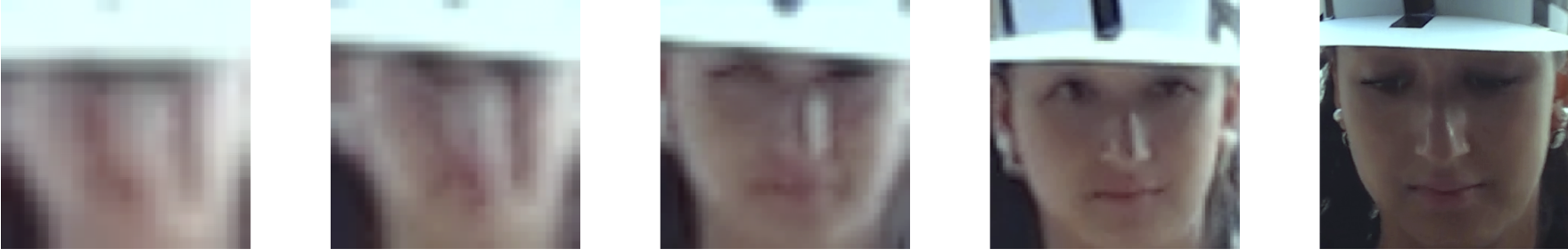} }}
        \addtocounter{subfigure}{-1}
        \qquad
        \subfloat[]{{\includegraphics[width=4.40cm]{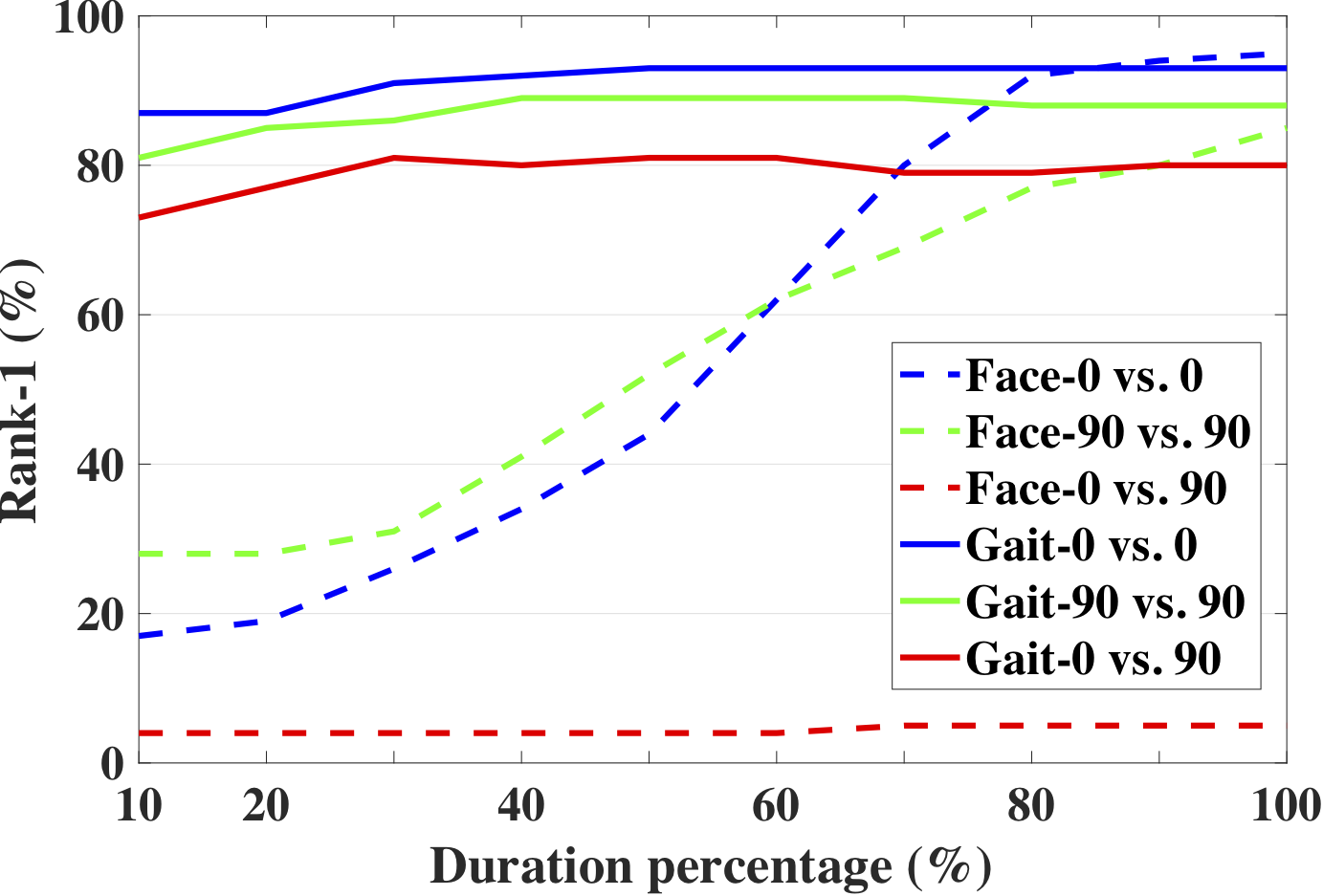} }}%
        \subfloat[]{{\includegraphics[width=4.40cm]{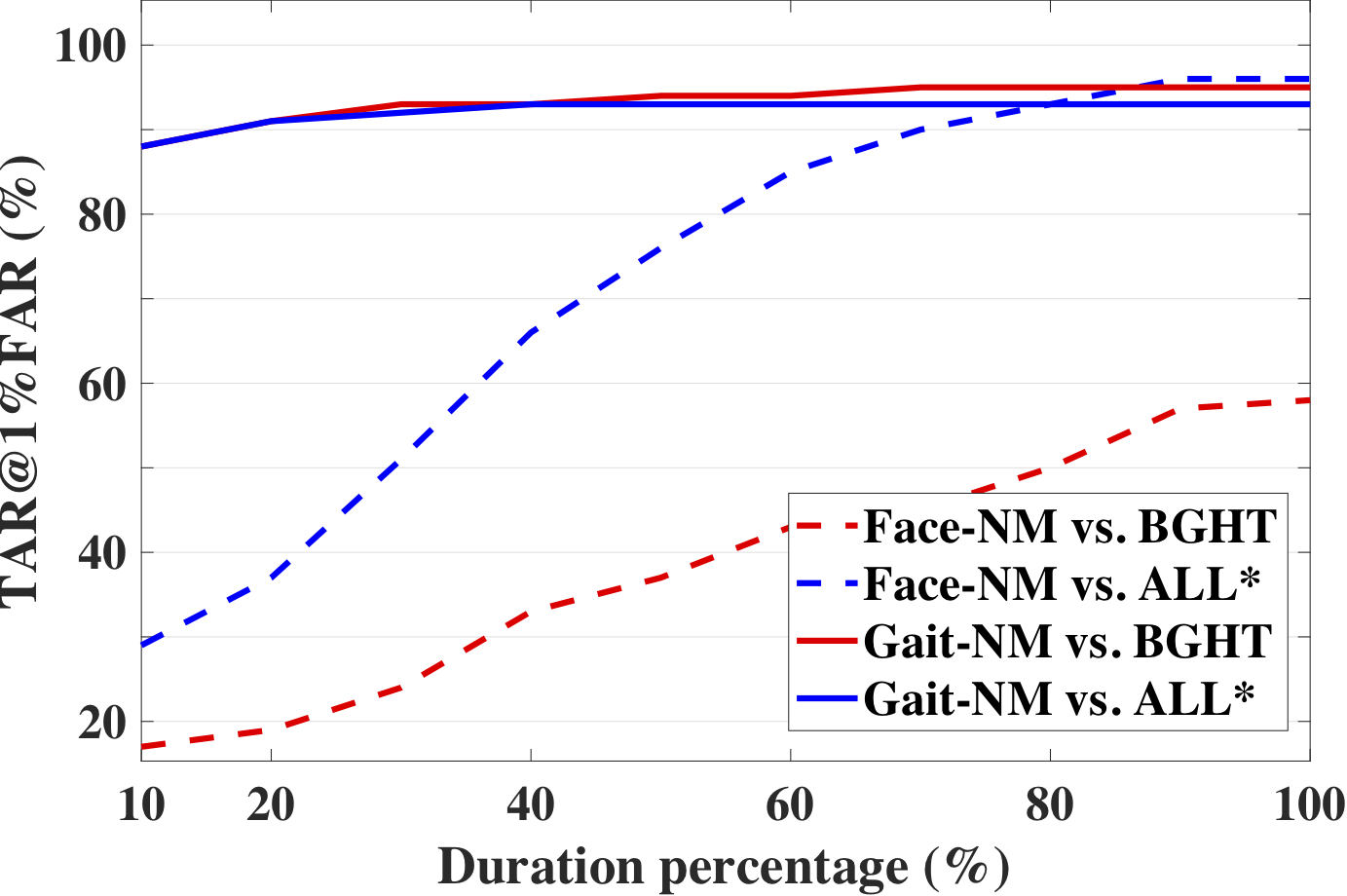} }}%
        \caption{Comparison of gait and face recognition on CASIA-B and FVG.
        Classification accuracy scores along with video duration percentage are calculated.
        (a) In CASIA-B, both gait and face recognition are performed in three scenarios: frontal-frontal ($0^\circ$ vs.~$0^\circ$), side-side ($90^\circ$ vs.~$90^\circ$) and frontal-side ($0^\circ$ vs.~$90^\circ$). (b) In FVG, both recognitions use NM vs.~BGHT and NM vs.~ALL* protocols. Detected face examples are shown on the top of each frontal and side view plots under various video duration percentage.}%
        \label{fig:face-gait}
    \end{figure}

    \subsubsection{Gait vs.~Face Recognition on CASIA-B}
    In this experiment, we select the videos of the NM as gallery and both CL and BG are probes.
    We compare gait and face recognition in three scenarios: frontal-frontal, side-side and side-frontal viewing angles.
    Fig.~\ref{fig:face-gait} shows the Rank-$1$ scores over the time duration.
    As the video begins, GaitNet is significantly superior to face in all scenarios since our $\fstagait$ can capture discriminative information such as body shape in low-resolution images, as mentioned in Sec.~\ref{sec:ablation_overtime}, while faces are of too low resolution to perform meaningful recognition.
    As time progresses, GaitNet is stable to the resolution change and view variations, with increasing accuracy.
    In comparison, face recognition always has lower accuracies throughout the entire duration, except the frontal-frontal view face recognition slightly outperforms gait in the last $20\%$ of the duration, which is expected as this is toward the ideal scenario for face recognition to shine.
    Unfortunately, for side-side or side-frontal views, face recognition continues to struggle even at the end of the duration.



    \subsubsection{Gait vs.~Face Recognition on FVG}
    We further compare GaitNet with ArcFace on FVG with NM-BGHT and NM-ALL* protocols.
    Note that the videos of NM-BGHT contain variations in carrying bags and wearing hat.
    The videos of ALL*, different from ALL in Tab.~\ref{tab:fvg}, include all the variations in FVG except carrying and wearing hat variations (refer to Tab.~\ref{tab:fvg_database} for details).
    As shown in Fig.~\ref{fig:face-gait}, on the BGHT protocol, gait outperforms face in the entire duration, since wearing hat dramatically affects face recognition but not gait recognition.
    For ALL* protocol, face outperforms gait in the last $20\%$ duration because by then low resolution is not an issue and FVG has frontal-view faces.



\begin{figure}[t]
        \includegraphics[width=8cm]{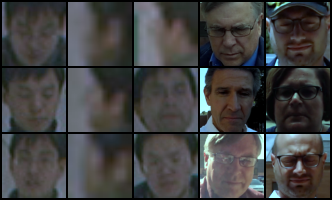}
        \centering
        \caption{Examples in CASIA-B and FVG where the SOTA face recognizer ArcFace fails.
        The first row is the image of probe set; the second row is the recognized wrong person in gallery; and the third row shows the genuine gallery.
        The first three columns are three scenarios of CASIA-B and the last two columns are two protocols of FVG.}
        \label{fig:face-wrong}
    \end{figure}
\begin{figure}[t]
        \includegraphics[width=6cm]{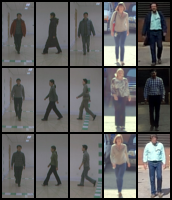}
        \centering
        \caption{Failure cases of GaitNet on CASIB-B and FVG due to blurry and illumination conditions.
        The rows and columns are defined the same as Fig.~\ref{fig:face-wrong}.}
        \label{fig:gait-wrong}
    \end{figure}

    Figure~\ref{fig:face-wrong} shows some examples in CASIB-B and FVG, which are incorrectly recognized by face recognition.
    We also show some images (video frames) for which our GaitNet fails to recognize in Fig.~\ref{fig:gait-wrong}.
    The low resolution and illumination conditions in these videos are the main reasons for failure.
    Note that while video-based alignment~\cite{liu2010video,tai2019towards} or super-resolution approaches~\cite{chen2018fsrnet} might help to enhance the image quality, their impact to recognition is beyond the scope of this work.


    \subsection{Runtime Speed}
    System efficiency is an essential metric for many vision systems including gait recognition.
    We calculate the efficiency while each of the $5$ gait recognition methods processing one video of FVG dataset on the same desktop with GeForce GTX 1080 Ti GPU.
    All the coding are implemented in PyTorch Framework of Python programming language. Parallel computing of batch processing is enabled for GPU on all the inference models, where batch size is number of samples in the probe. Alphapose and Mask-R-CNN takes batch size of $1$ as input in inference.
    As shown in Tab.~\ref{tab:runtime}, our method is faster than the pose estimation method because of 1) an accurate, yet slow, version of AlphaPose~\cite{fang2017rmpe} is required for model-based gait recognition method; 2) only low-resolution input of $32\times64$ pixels is needed for GaitNet.
    Further, our method has similar efficiency as the recent CNN-based gait recognition methods.

    \begin{table}[t!]
        \centering
        \caption{Runtime (ms per frame) comparison on FVG dataset.}
        \label{tab:runtime}
        \resizebox{0.75\linewidth}{!}{%
        \begin{tabular}{lccc}
            \toprule
            Methods & Pre-processing & Inference & Total \\  \midrule
            PE-LSTM & $224.4$    &      $0.1$ &    $224.5$      \\
            GEINet~\cite{shiraga2016geinet} & $89.5$    &      $1.5$  & $91.0$     \\
            DCNN~\cite{alotaibi2017improved} & $89.5$    &      $1.7$  & $91.2$     \\
            LB~\cite{wu2017comprehensive} & $89.5$    &      $1.3$ & $90.8$      \\
            GaitNet (ours)                   & $89.5$    &       $1.0$  & $90.5$     \\
            \bottomrule
        \end{tabular}}
    \end{table}

    \section{Conclusion}
    This paper presents an autoencoder-based method termed GaitNet that can disentangle appearance and gait feature representation from raw RGB frames, and utilize a multi-layer LSTM structure to further leverage temporal information to generate a gait representation for each video sequence.
    We compare our method extensively with the state of the arts on CASIA-B, USF, and our collected FVG datasets.
    The superior results show the generalization and promise of the proposed feature disentanglement approach.
    We hope that in the future, this disentanglement approach is a viable option for other vision problems where motion dynamics needs to be extracted while being invariant to confounding factors, \emph{e.g.}, expression recognition with invariance to facial appearance, activity recognition with invariance to clothing.


    %

    %

    \ifCLASSOPTIONcompsoc
    \section*{Acknowledgments}
    \else
    \section*{Acknowledgment}
    \fi

    This work was partially sponsored by the Ford-MSU Alliance program, and
    the Army Research Office under Grant Number W911NF-18-1-0330. The views and conclusions contained in this
    document are those of the authors and should not be interpreted
    as representing the official policies, either expressed
    or implied, of the Army Research Office or the U.S.~Government.
    The U.S.~Government is authorized to reproduce
    and distribute reprints for Government purposes notwithstanding
    any copyright notation herein.

    \ifCLASSOPTIONcaptionsoff
    \newpage
    \fi



    %

    {\small
    \bibliographystyle{IEEEtran}
    \bibliography{egbib}
    }

    %
    %



    \begin{IEEEbiography}[{\includegraphics[width=1in,height=1.25in,clip,keepaspectratio]{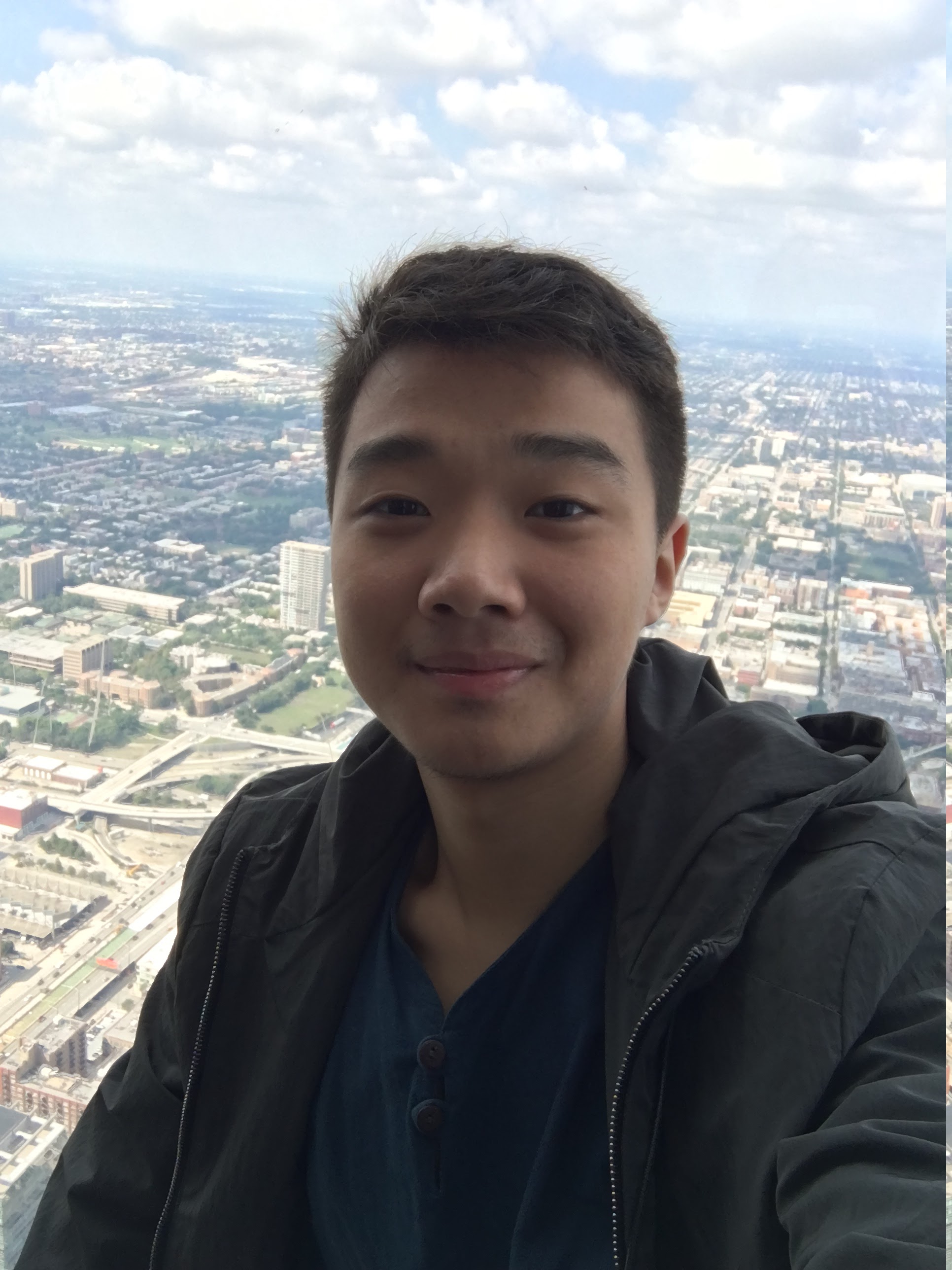}}]{Ziyuan Zhang}
        is now pursuing his B.S. in Computer Science from Michigan State University.
        His research areas of interest include deep learning and computer vision.
    \end{IEEEbiography}
    \begin{IEEEbiography}[{\includegraphics[width=1in,height=1.25in,clip,keepaspectratio]{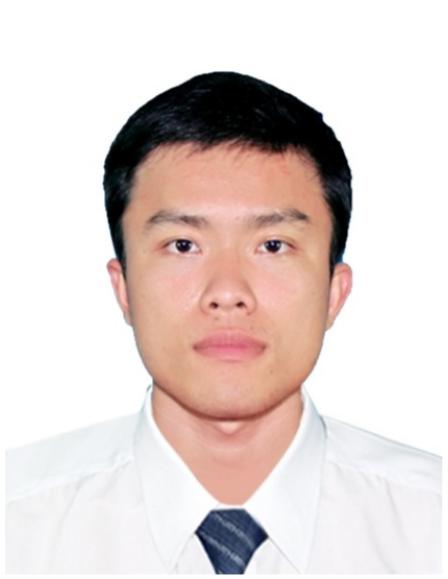}}]{Luan Tran}
        received his B.S. in Computer Science from Michigan State University with High Honors in $2015$. He is now pursuing his Ph.D. also at Michigan State University in the area of deep learning and computer vision. His research areas of interest include deep learning and computer vision, in particular, face modeling and face recognition.
    \end{IEEEbiography}
    \begin{IEEEbiography}[{\includegraphics[width=1in,height=1.25in,clip,keepaspectratio]{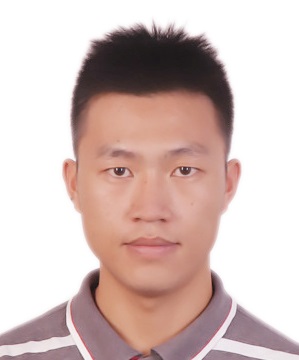}}]{Feng Liu}
        is currently a post-doc researcher in the Computer Vision Lab at Michigan State University. He received the Ph.D. degree in Computer Science from Sichuan University, China in $2018$. His main research interests focus on computer vision and pattern recognition, specifically for $3$D modeling, $2$D and $3$D face recognition.
    \end{IEEEbiography}
    \begin{IEEEbiography}[{\includegraphics[width=1in,height=1.25in,clip,keepaspectratio]{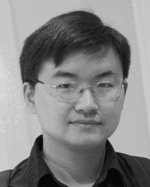}}]{Xiaoming Liu}
        is an Associate Professor at the Department of Computer Science and Engineering of Michigan State University. He received the Ph.D. degree in Electrical and Computer Engineering from Carnegie Mellon University in 2004. Before joining MSU in Fall $2012$, he was a research scientist at General Electric (GE) Global Research. His research interests include computer vision, machine learning, and biometrics. As a co-author, he is a recipient of Best Industry Related Paper Award runner-up at ICPR $2014$, Best Student Paper Award at WACV $2012$ and $2014$, and Best Poster Award at BMVC $2015$. He has been the Area Chair for numerous conferences, including FG, ICPR, WACV, ICIP, CVPR, ICCV, and ICLR. He is the program chair of WACV $2018$, and BTAS $2018$. He is an Associate Editor of Neurocomputing journal, Pattern Recognition Letters, and Pattern Recognition. He has authored more than $150$ scientific publications, and has filed $22$ U.S. patents.
    \end{IEEEbiography}




\end{document}